\documentclass[journal, 10pt]{IEEEtran}

\usepackage{amsmath,amsfonts,amssymb,amsbsy}
\usepackage{amsthm}

\usepackage{caption}
\usepackage{subcaption}
\usepackage{epstopdf}
\usepackage{amsfonts,epstopdf, tikz, pgfplots}
\usepackage{float}

\usepackage{multirow}
\usepackage{verbatim}
\usepackage{pslatex}
\usepackage{cite,url,bm}
\usepackage{tablefootnote, threeparttable, booktabs, multirow}
\usepackage{algorithmicx}
\usepackage[ruled]{algorithm}
\usepackage{algpseudocode}

\alglanguage{pseudocode}
\usepackage{epsf,psfig}
\usepackage{latexsym,epsfig}
\usepackage{multirow}
\usepackage{xcolor}

\usepackage{graphicx}
\usepackage{tikz}
\usepackage{epstopdf}
\usepackage{verbatim}

\usetikzlibrary{arrows,decorations.pathmorphing,backgrounds,positioning,fit,petri}

\usetikzlibrary{shapes,arrows}

 \pdfoutput=1
\usepackage[english]{babel}
\makeatletter
\adddialect\l@ENGLISH\l@english
\makeatother

\pgfplotsset{footnotesize}
\usepackage{parskip}
\usepackage[T1]{fontenc}
\usepackage{pgf}
\usepackage[normalem]{ulem}

\usepackage[inline]{enumitem}
\usepackage[autoplay]{animate}

\usepackage{subcaption}
\usepackage{float}
\definecolor{colorsrc}{rgb}{0.36, 0.54, 0.66}

\definecolor{colorwnd2}{rgb}{0.91, 0.84, 0.42}
\definecolor{colorwnd}{rgb}{0.8, 0.0, 0.1}
\definecolor{colorfdd}{rgb}{0.44, 0.16, 0.39}
\definecolor{colorshi}{rgb}{0.55, 0.71, 0.0}
\definecolor{colornan}{rgb}{0.8, .33, 0}
\definecolor{colornan}{rgb}{0.72, 0.53, 0.04}
\definecolor{darkcyan}{rgb}{0.0, 0.55, 0.55}
\definecolor{colordfd}{rgb}{0.0, 0.2, 1.0}
\definecolor{colorlck}{rgb}{0.0, 0.9, 0.9}
\definecolor{pinegreen}{rgb}{0.0, 0.47, 0.44}

\newcommand{\mat}[1]{\mathbf{#1}}

\newcommand{\norm}[1]{\left\|#1\right\|}

\newcommand{\tb}{\textbf}

\def\bmt{\left[\begin{matrix}}

\def\emt{\end{matrix}\right]}

\def\tb{\textbf}

\def\mc {\mathcal}

\def\bd{\mathbf{d}}
\def\be{\mathbf{e}}
\def\fb{\mathbf{f}}

\def\bm{\mathbf{m}}

\def\bs{\mathbf{s}}
\def\bu{\mathbf{u}}
\def\by{\mathbf{y}}
\def\and{\text{~and~}}
\def\barN{\bar{N}}

\def\trace{\textrm{trace}}
\def\etal{\textit{et al.}}
\def\R{\mathbb{R}}
\def\bA{\mathbf{A}}
\def\bB{\mathbf{B}}
\def\bX{\mathbf{Y}}

\def\bY{\mat{Y}}
\def\bD{\mathbf{D}}
\def\bQ{\mathbf{Q}}
\def\bM{\mathbf{M}}
\def\bS{\mathbf{S}}
\def\bE{\mathbf{E}}
\def\Fb{\mathbf{F}}

\def\barX{\bar{\mathbf{Y}}}

\def\barS{\bar{\mathbf{S}}}

\def\bW{\mathbf{W}}

\def\ben{\begin{equation*}}
\def\een{\end{equation*}}
\def\beaa{\begin{eqnarray*}}
\def\eeaa{\end{eqnarray*}}
\def\bea{\begin{eqnarray}}
\def\eea{\end{eqnarray}}
\usepackage{lipsum}
\newcommand{\subparagraph}{}
\usepackage{titlesec}
\usepackage[english]{babel}
\usepackage{fancyhdr}

\titlespacing\section{0pt}{2pt plus 1pt minus 1pt}{0pt plus 2pt minus 2pt}
\titlespacing\subsection{0pt}{2pt plus 1pt minus 1pt}{0pt plus 0pt minus 2pt}
\begin{document}
\title{Histopathological Image Classification using Discriminative Feature-oriented Dictionary Learning}
\author{Tiep Huu Vu$^{\dagger}$,~\IEEEmembership{Student Member,~IEEE,} Hojjat Seyed Mousavi$^{\dagger}$,~\IEEEmembership{Student Member,~IEEE,} \\ Vishal Monga$^{\dagger}$,~\IEEEmembership{Senior Member,~IEEE,}   Ganesh Rao$^{*}$ and UK Arvind Rao$^{*}$ \vspace{-0.1in}
\thanks{$^{\dagger}$T. H. Vu, H. S. Mousavi, and V. Monga are with the Department of Electrical Engineering, Pennsylvania State University, University Park, PA 16802, USA (e-mail: thv102@psu.edu).
\par
$^{*}$Ganesh Rao is with the Department of Neurosurgery, and UK Arvind Rao is with the Department of Bionformatics and Computational Biology, both at the University of Texas MD Anderson  Cancer Center, Houston, TX, USA.
\par
Reasearch was supported by the Army Research Office (ARO) grant number W911NF-14-1-0421 (to V.M.), National Institute of Health NIH grant KNS070928 (to G.R.), NCI Cancer Center Support Grant NCI P30 CA016672 and Career Development Award from the Brain Tumor SPORE (to A.R.).}} 
\maketitle
\thispagestyle{fancy}
\pagestyle{empty}		
\vspace{-1in}
\begin{abstract}
\label{abstract}
In histopathological image analysis, feature extraction for classification is a challenging task due to the diversity of histology features suitable for each problem as well as presence of rich geometrical structures. In this paper, we propose an automatic feature discovery framework via learning class-specific dictionaries and present a low-complexity method for classification and disease grading in histopathology. Essentially, our Discriminative Feature-oriented Dictionary Learning (DFDL) method learns class-specific dictionaries such that under a sparsity constraint, the learned dictionaries allow representing a new image sample parsimoniously via the dictionary corresponding to the class identity of the sample. At the same time, the dictionary is designed to be poorly capable of representing samples from other classes. Experiments on three challenging real-world image databases: 1) histopathological images of intraductal breast lesions, 2) mammalian kidney, lung and spleen images provided by the Animal Diagnostics Lab (ADL) at Pennsylvania State University, and 3) brain tumor images from The Cancer Genome Atlas (TCGA) database, reveal the merits of our proposal over state-of-the-art alternatives. {Moreover, we demonstrate that DFDL exhibits a more graceful decay in classification accuracy against the number of training images which is highly desirable in practice where generous training is often not available}.
\end{abstract}
\textbf{\small \textit{  Index terms---}Histopathological image classification, Sparse coding, Dictionary learning, Feature extraction, Cancer grading.}
\section{Introduction}
\label{sec:intro}
 Automated histopathological image analysis has recently become a significant research problem in medical imaging and there is an increasing need for developing quantitative image analysis methods as a complement to the effort of pathologists in diagnosis process. Consequently, an emerging class of problems in medical imaging focuses on the the development of computerized frameworks to classify histopathological images \cite{Gurcan2009,Srinivas2013,Srinivas2014SHIRC,Nandita2013,Mousavi2015JPI}. These advanced image analysis methods have been developed with {three main purposes of (i) relieving the workload on pathologists by sieving out obviously diseased and also healthy cases, which allows specialists to spend more time on more sophisticated cases; (ii) {reducing inter-expert variability}; and (iii) understanding the underlying reasons for a specific diagnosis that pathologists might not realize.}
\par
In the diagnosis process, pathologists often look for problem-specific visual cues, or features, in histopathological images in order to categorize a tissue image as one of the possible categories. These features might come from the distinguishable characteristics of cells or nuclei, for example, size, shape or texture\cite{Dundar2011,Gurcan2009}. They could also come from spatially related structures of cells\cite{Mousavi2015JPI, tosun2011graph,Srinivas2014SHIRC,doyle2008automated}. In some cancer grading problems, features might include the presence of particular regions \cite{Mousavi2015JPI, hou2015efficient}. Consequently, different customized feature extraction techniques for a variety of problems have been developed based on these observed features\cite{Orlov2008,Shamir2008,Gultekin2014,Shi2014,Minaee2013prediction}. Morphological image features have been utilized in medical image segmentation\cite{zana2001segmentation} for detection of vessel-like patterns. Wavelet features and histograms are also a popular choice of features for medical imaging\cite{chapelle1999support,unser2003guest}. Graph-based features such as Delaunay triangulation, Vonoroi diagram, minimum spanning tree\cite{doyle2008automated}, query graphs\cite{ozdemir2013hybrid} have been also used to exploit spatial structures. Orlov \etal \cite{Orlov2008,Shamir2008} have proposed a multi-purpose framework that collects texture information, image statistics and transforms domain coefficients to be set of features. For classification purposes, these features are combined with powerful classifiers such as neural networks or support vector machines (SVMs). Gurcan \etal \cite{Gurcan2009} provided detailed discussion of feature and classifier selection for histopathological analysis.
\par Sparse representation frameworks have also been proposed for medical applications recently \cite{Srinivas2014SHIRC, Nandita2013, kopriva2015offset}. Specifically, Srinivas \etal \cite{Srinivas2013,Srinivas2014SHIRC} presented a multi-channel histopathological image as a sparse linear combination of training examples under channel-wise constraints and proposed a residual-based classification technique. Yu \etal\cite{Yu2011} proposed a method for cervigram segmentation based on sparsity and group clustering priors. Song \etal\cite{song2015locality,song2015large}  proposed a locality-constrained and a  large-margin representation method for medical image classification. In addition, Parvin \etal \cite{Nandita2013} combined a dictionary learning framework with an autoencoder to learn sparse features for classification. Chang \etal \cite{chang2013characterization} extended this work by adding a spatial pyramid matching to enhance the performance.
\subsection{Challenges and Motivation} 
\label{sub:motivation}
While histopathological analysis shares some traits with other image classification problems, there are also principally distinct challenges specific to histopathology.  The central challenge comes from the geometric richness of tissue images, resulting in the difficulty of obtaining reliable discriminative features for classification. Tissues from different organs have structural and morphological diversity which often leads to highly customized feature extraction solutions for each problem and hence the techniques lack broad applicability.
\par Our work aims to produce a more versatile histopathological image classification system through the design of discriminative, class-specific dictionaries which is hence capable of automatic feature discovery using example training image samples. Our proposal evolves from the sparse representation-based classifier (SRC)\cite{Wright2009SRC} which has received significant attention recently\cite{huang2006sparse,Bahrampour2014,Mousavi2014ICIP}. Wright \etal\cite{Wright2009SRC} proposed SRC with the assumption that given a sufficient collection of training samples from one class, which is referred as a dictionary, any other test sample from the same class can be roughly expressed as a linear combination of these training samples. As a result, any test sample has a \textit{sparse} representation in terms of a big dictionary comprising of class-specific sub-dictionaries. Recent work has shown that learned and data adaptive dictionaries significantly outperform ones constructed by simply stacking training samples together as in \cite{Wright2009SRC}. In particular, methods with class-specific constraints\cite{zhang2010discriminative,Zhuolin2013LCKSVD,Suo2014,Meng2011FDDL} are known to further enhance classification performance.

\par
\par

\par
\begin{figure}[t]
\centering
  \includegraphics[width=0.47\textwidth]{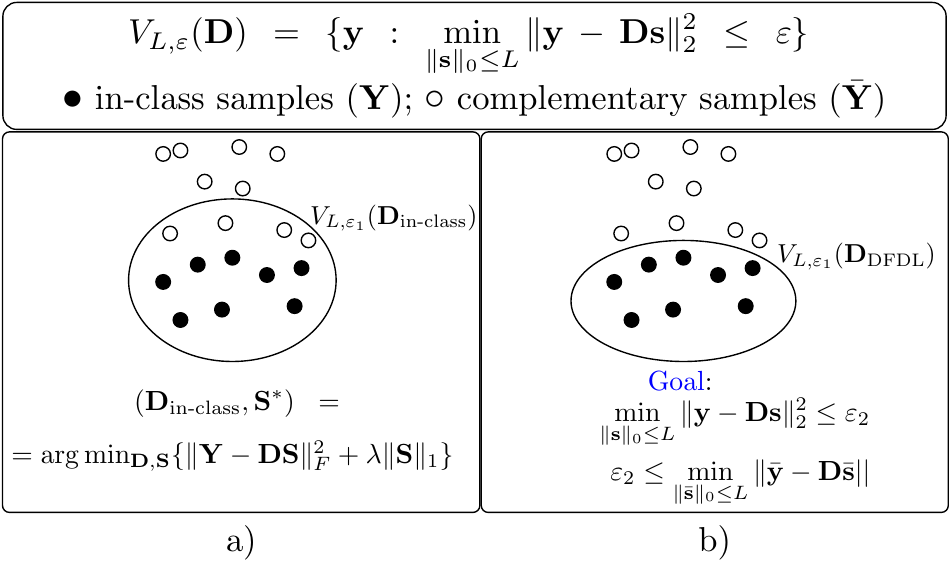}
  \vspace{-0.1in}
  \caption{\small Main idea: a) The sparse representation space of learned dictionary using in-class samples only, e.g. KSVD\cite{Aharon2006KSVD} or ODL\cite{mairal2010online}($V_{L,\varepsilon_1}(\bD_{\text{in-class}})$ may also cover some complementary samples), and b) desired DFDL ($V_{L,\varepsilon_2}(\bD_{\text{DFDL}})$ cover in-class samples only).}
  \label{fig: idea}
\end{figure}

\pagestyle{headings}
Being mindful of the aforementioned challenges, we design via optimization, a discriminative dictionary for each class by imposing sparsity constraints that minimizes intra-class differences, {while simultaneously} emphasizing inter-class differences. On one hand, small intra-class differences encourage the comprehensibility of the set of learned bases, which has ability of representing in-class samples with only few bases (intra class sparsity). This encouragement forces the model to find the representative bases in that class. On the other hand, large inter-class differences prevent bases of a class from sparsely representing samples from other classes.
Concretely, given a dictionary from a particular class $\bD$ with $k$ bases and a certain sparsity level $L\ll k$, we define an \emph{$L$-subspace} of $\bD$ as a span of a subset of $L$ bases from $\bD$. Our proposed Discriminative Feature-oriented Dictionary Learning (DFDL) aims to build dictionaries with this key property: any sample from a class is reasonably close to \emph{an} $L$-subspace of the associated dictionary while a complementary sample is far from \emph{any} $L$-subspace of that dictionary. Illustration of the proposed idea is shown in Fig. \ref{fig: idea}.

\subsection{Contributions}
\label{sub:our_contri}

The main contributions of this paper are as follows:
	\par 1) \textbf{A new discriminative dictionary learning method\footnote{ The preliminary version of this work was presented at IEEE International Symposium on Biomedical Imaging, 2015\cite{vu2015dfdl}.}} for automatic feature discovery in histopathological images is presented to mitigate the generally difficult problem of feature extraction in histopathological images. Our \emph{discriminative} framework learns dictionaries that emphasize inter-class differences while keeping intra-class differences small, resulting in {enhanced} classification performance. The design is based on solving a sparsity constrained optimization problem, for which we develop a tractable algorithmic solution.
	\par
\begin{figure}[t]
\begin{center}
\begin{tikzpicture}
  \node (u1) [inner sep = 0pt] {\includegraphics[width = 0.23\textwidth]{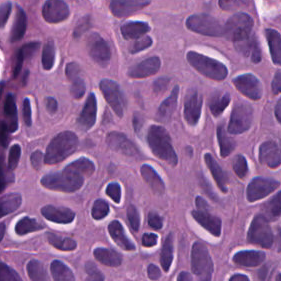}};
  \node (u2) [inner sep = 0pt, anchor = west, right = 5pt of u1.east] {\includegraphics[width = 0.23\textwidth]{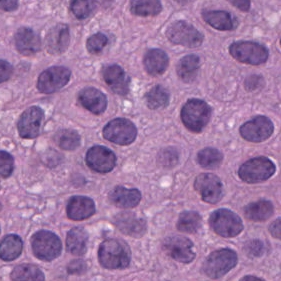}};
\end{tikzpicture}
\caption{\small Samples form IBL dataset: left-UDH, right-DCIS}
\label{fig:iblsamples}
\end{center}
\end{figure}

\begin{figure}
\begin{center}
\begin{tikzpicture}
  \node (u1) [inner sep = 0pt] {\includegraphics[width = 0.23\textwidth]{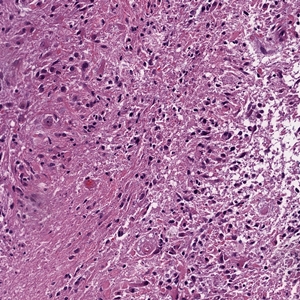}};
  \node (u2) [inner sep = 0pt, anchor = west, right = 5pt of u1.east] {\includegraphics[width = 0.23\textwidth]{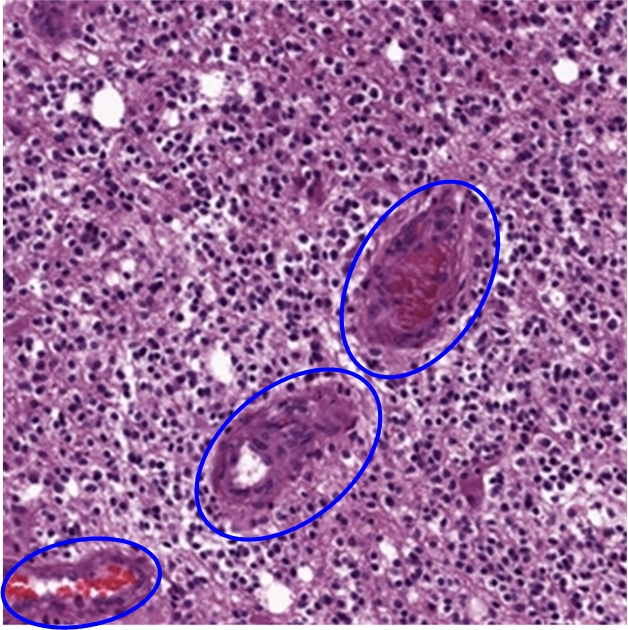}};


\end{tikzpicture}
\caption{\small Samples form TCGA dataset. Left: regions without MVP. Right: regions with MVP are inside blue ovals.}
\label{fig:tcgasamples}
\end{center}
\end{figure}

\begin{figure}
\begin{center}
\begin{tikzpicture}
  \node (u1) [inner sep = 0pt] {\includegraphics[width = 125pt]{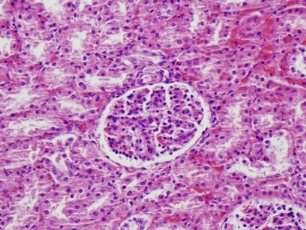}};
  \node (u2) [inner sep = 0pt, anchor = west, right = 2pt of u1.east] {\includegraphics[width = 125pt]{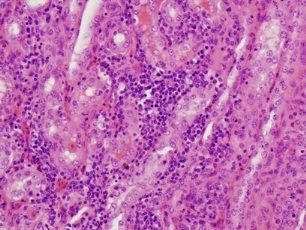}};

  \node (d1) [inner sep = 0pt, anchor = north, below = 5pt of u1.south] {\includegraphics[width = 125pt]{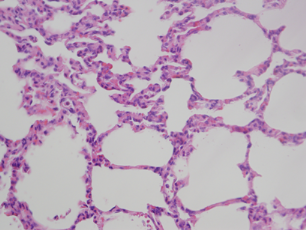}};
  \node (d2) [inner sep = 0pt, anchor = west, right = 2pt of d1.east] {\includegraphics[width = 125pt]{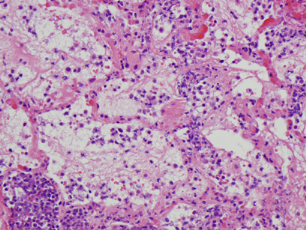}};

  \node (d1) [inner sep = 0pt, anchor = north, below = 5pt of d1.south] {\includegraphics[width = 125pt]{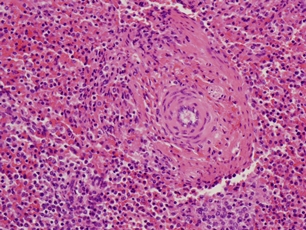}};
  \node (d2) [inner sep = 0pt, anchor = west, right = 2pt of d1.east] {\includegraphics[width = 125pt]{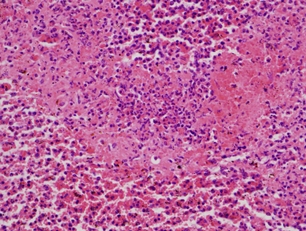}};

\end{tikzpicture}
\caption{\small Samples form ADL dataset. First row: kidney. Second row: lung. Last row: spleen. Left: healthy. Right: inflammatory.}
\label{fig:adlsamples}
\end{center}
\end{figure}

 2) \textbf{Broad Experimental Validation and Insights}. Experimental validation of DFDL is carried out on three diverse histopathological datasets to show its broad applicability. \textit{The first dataset} is courtesy of the Clarian Pathology Lab and Computer and Information Science Dept., Indiana University-Purdue University Indianapolis (IUPUI). The images acquired by the process described in \cite{Dundar2011} correspond to human Intraductal Breast Lesions (IBL). Two well-defined categories will be classified: Usual Ductal Hyperplasia (UDH)--benign, and Ductal Carcinoma In Situ (DCIS)--actionable. \textit{The second dataset} contains images of brain cancer {(glioblastoma or GBM)} obtaind from The Cancer Genome Atlas (TCGA) \cite{TCGA} provided by the National Institute of Health, and will henceforth be referred as the TCGA dataset. For this dataset, we address the problem of detecting MicroVascular Proliferation (MVP) regions, which is an important indicator of a high grade glioma (HGG)\cite{Mousavi2015JPI}. \textit{The third dataset} is provided by the Animal Diagnostics Lab (ADL), The Pennsylvania State University. It contains tissue images from three mammalian organs - kidney, lung and spleen. For each organ, images will be assigned into one of two categories--healthy or inflammatory.
	The samples of these three datasets are given in Figs.\ \ref{fig:iblsamples},  \ref{fig:tcgasamples}, and \ref{fig:adlsamples}, respectively. Extensive experimental results show that our method outperforms many competing methods, particularly in low training scenarios. In addition, Receiver Operating Characteristic (ROC) curves are provided that facilitate a trade-off between false alarm and miss rates.

\par 3) \textbf{Complexity analysis.} We derive the computational complexity of DFDL as well as competing dictionary learning methods in terms of approximate number of operations needed. We also report experimental running time of DFDL and three other dictionary learning methods.
\par 4) \textbf{Reproducibility}. All results in the manuscript are reproducible via a user-friendly software\footnote{The software can be downloaded at \url{http://signal.ee.psu.edu/dfdl.html}}. The software (MATLAB toolbox) is also provided with the hope of usage in future research and comparisons via peer researchers.

\par

{The remainder of this paper is organized as follows. Our proposed DFDL via a sparsity constrained optimization and the solution for the said optimization problem are detailed in Section \ref{sec:contributions}. Section \ref{sec:overallclassification} also presents our algorithmic classification procedures for the three diverse histopathological problems stated above. Section \ref{sec:experiment_results} presents classification accuracy as well as run-time complexity comparisons with existing methods in the literature to reveal merits of the proposed DFDL. A detailed analytical comparison of complexity against competing dictionary learning methods is provided in the Appendix. Section \ref{sec:Conclusion} concludes the paper.}

\section{Contributions}
\label{sec:contributions}
\subsection{Notation} 
\vspace{-0.05in}
\label{sub:notaions}
The vectorization of a small block (or patch)\footnote{In our work, a training vector is obtained by vectorizing all three RGB channels followed by concatenating them together to have a long vector.} extracted from an image is denoted as a column vector $\by \in \R^d$ which will be referred as a sample.
In a classification problem where we have $c$ different categories, collection of all data samples from class $i$ ($i$ can vary between $1$ to $c$) forms the matrix $\bX_i \in \R^{d\times N_i}$ and let $\bar{\bX}_i \in \R^{d \times \bar{N}_i}$ be the matrix containing all complementary data samples i.e. those that are not in class $i$. We denote by $\bD_i \in \R^{d \times k_i}$ the dictionary of class $i$ that is desired to be learned through our DFDL method. 
\par
{For a vector $\bs \in \R^k$, we denote by $\|\bs\|_0$ the number of its non-zero elements. The sparsity constraint of $\bs$ can be formulated as $\|\bs\|_0 \le L$. For a matrix $\bS$ , $\|\bS\|_0 \le L$ means that \emph{each} column of $\bS$ has no more than $L$ non-zero elements.}
\subsection{Discriminative Feature-oriented Dictionary Learning} 
\label{sec:DFDL}
We aim to build {class-specific} dictionaries $\bD_i$ such that each $\bD_i$ can sparsely represent samples from class $i$ but is \emph{poorly} capable of representing its complementary samples with small number of bases. Concretely, for the learned dictionaries we need:
\begin{eqnarray*}
  &\displaystyle \min_{\|\bs_l\|_0 \leq L_i }\|\by_l - \bD_i \bs_l\|_2^2,~~ \forall l = 1, 2, \dots, N_i & \text{to be small}\\
 \text{and } &\displaystyle \min_{\|\bar{\bs}_m\|_0 \leq L_i}\|\bar{\by}_m - \bD_i \bar{\bs}_m\|_2^2,~~ \forall m = 1,2,\dots, \bar{N}_i & \text{to be large.}
\end{eqnarray*}
 where $L_i$ controls the sparsity level. These two sets of conditions could be simplified in the matrix form:
 \begin{eqnarray}
  \label{eqn:intra}
   \text{intra-class differences: }\displaystyle\frac{ 1  } { N_i } \min_{   \|\bS_i\|_0 \leq L_i }\|\bX_i - \bD_i \bS_i\|_F^2 & \text{small,}\\
  \label{eqn:inter}
 \text{inter-class differences: } \displaystyle \frac{ 1  } { \bar{N_i} } \min_{\|\bar{\bS}_i\|_0 \leq L_i}\|\bar{\bX}_i - \bD_i \bar{\bS}_i\|_F^2 & \text{large.}
 \end{eqnarray}

The averaging operations $\left(\displaystyle\frac{ 1  } { N_i } \text{~and~} \displaystyle\frac{1}{\bar{N_i}}\right)$  are taken here  for avoiding the case where the largeness of inter-class differences is solely resulting from $\bar{N_i} \gg N_i$.
 \par For simplicity, from now on, we consider only one class and drop the class index in each notion, i.e., using $\bX, \bD, \bS, \bar{\bS}, N, \bar{N}, L$ instead of $\bX_i, \bD_i, \bS_i, \bar{\bS}_i, N_i, \bar{N}_i$ and $L_i$.
 Based on the argument above, we formulate the optimization problem for each dictionary:
 \begin{equation}
   \bD^*= \arg \min _{\bD} \Big(\frac{ 1  } { N }\min_{\|\bS\|_0 \le L}\|\bX - \bD \bS\|_F^2 - \frac{ \rho  } { \bar{N} } \min_{\|\bar{\bS}\|_0 \le L}\|\bar{\bX} - \bD \bar{\bS}\|_F^2 \Big),
 \label{eqn:findDopt}
 \end{equation}
 where $\rho$ is a positive regularization parameter. The first term in the above optimization problem encourages intra-class differences to be small, while the second term, with minus sign, emphasizes inter-class differences. By solving the above problem, we can jointly find the appropriate dictionaries as we desire in (\ref{eqn:intra}) and (\ref{eqn:inter}).
 \par
 \textbf{How to choose $L$:} The sparsity level $L$ for classes might be different. For one class, if $L$ is too small, the dictionary might not appropriately express in-class samples, while if it is too large, the dictionary might be able to represent complementary samples as well. In both cases, the classifier might fail to determine identity of one new test sample. We propose a method for estimating $L$ as follows. First, a dictionary is learned using ODL\cite{mairal2010online} using in-class samples $\bY$ only:
 \begin{equation}
     (\bD^0, \bS^0) = \arg\min_{\bD, \bS}\{\|\mathbf{Y} - \bD \bS\|_F^2 + \lambda\|\bS\|_1\},
     \label{eqn:findD0}
 \end{equation}
 where $\lambda$ is a positive regularization parameter controlling the sparsity level. {Note that the same $\lambda$ can still lead to different $L$ for different classes, depending on the intra-class variablity of each class. Without prior knowledge of those variablities, we choose the same $\lambda$ for every class.} After $\bD^0$ and $\bS^0$ have been computed, $\bD^0$ could be utilized as a warm initialization of $\bD$ in our algorithm, $\bS^0$ could be used to estimate the sparsity level $L$:
 \begin{equation}
    L \approx \frac{1}{N}\sum_{i=1}^{N}\|\bs_i^0\|_0.
     \label{eqn:findL}
 \end{equation}

\textbf{Classification scheme:} In the same manner with SRC \cite{Wright2009SRC}, a new patch $\by$ is classified as follows. Firstly, the sparse codes $\hat{\bs}$ are calculated via $l_1$-norm minimization:
\begin{equation}
    \hat{\bs} = \arg \min_{\bs} \big\{ \|\by - \bD_{total}\bs\|_2^2 + \gamma \|\bs\|_1 \big\},
    \label{eqn:class1}
\end{equation}
    where $\bD_{total} = [\bD_1, \bD_2, \dots, \bD_c]$ is the collection of all dictionaries and $\gamma$ is a scalar constant.
    Secondly, the identity of $\by$ is determined as: $\displaystyle \arg \min_{i \in \{1, \dots, c\}}\{r_i(\by)\} $ where
\begin{equation}
  r_i(\by) = \|\by-\bD_i \delta_i(\hat{\bs})\|_2
  \label{eqn:class2}
\end{equation}
    and $\delta_i(\hat{\bs})$ is part of $\hat{\bs}$ associated with class $i$.


\begin{figure*}[t]
\centering
  \includegraphics[width=\textwidth]{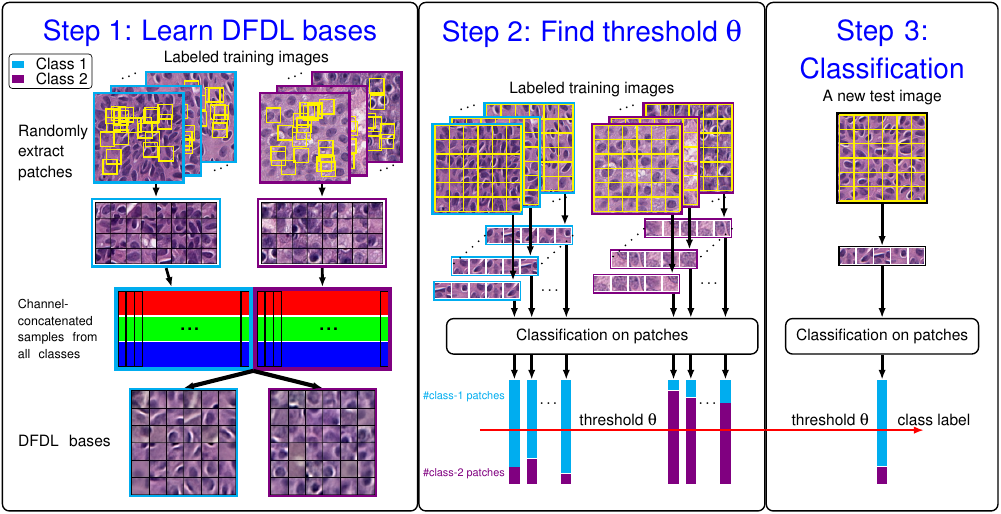}
  \vspace{-0.1in}
 \caption{\small IBL/ADL classification procedure }
  \label{fig: ibladlprocedure}
\end{figure*}
\subsection{Proposed solution}
\label{subsec: solution}
We use an iterative method to find the optimal solution for the problem in (\ref{eqn:findDopt}). Specifically, the process is iterative by fixing $\bD$ while optimizing $\bS$ and $\bar{\bS}$ and vice versa. 
\par
In the sparse coding step, with fixed $\bD$, optimal sparse codes $\bS^*, \bar{\bS}^*$ can be found by solving:
$$
       \bS^* = \arg \min_{\|\bS\|_0 \le L}\|\bX - \bD\bS\|_F^2;~~
       \bar{\bS}^* = \arg \min_{\|\bar{\bS}\|_0 \le L} \|\bar{\bX} - \bD\bar{\bS}\|_F^2.
       $$
\par  With the same dictionary $\bD$, these two sparse coding problems can be combined into the following one:
  \begin{equation}\label{eqn:findS}
    \hat{\bS}^* = \arg\min_{\|\hat{\bS}\|_0 \le L}\norm{\hat{\bX} - \bD\hat{\bS}}_F^2.
  \end{equation}
  with $\hat{\bX} = [\bX, \bar{\bX}]$ being the matrix of all training samples and $\hat{\bS} = [\bS, \bar{\bS}]$. This sparse coding problem can be solved effectively by OMP\cite{tropp2007signal} using SPAMS toolbox\cite{SPAMS}.

  \par
For the bases update stage, $\bD^*$ is found by solving: 
\begin{align}
  \label{subeqn:findD1}
  \bD^* &=& \arg\min_{\bD} \Big\{ \frac{ 1  } { N } \|\bX - \bD\bS\|_F^2 - \frac{ \rho } { \barN } \|\bar{\bX} - \bD\bar{\bS}\|_F^2 \Big\}, \\
  \label{subeqn:findD2}
  &=& \arg\min_{\bD} \big\{-2\trace(\bE \bD^\top) + \trace(\bD \Fb \bD^\top) \big\}.
\end{align}

\noindent
We have used the equation $\|\bM\|_F^2 = \trace(\bM\bM^\top)$ for any matrix $\bM$ to derive (\ref{subeqn:findD2}) from (\ref{subeqn:findD1}) and denoted:
\begin{equation}
\label{eqn:defineEF}
        \bE = \frac{1}{N} \bY \bS^\top - \frac{ \rho } { \barN } \barX\barS^\top; \quad
        \Fb = \frac{ 1  } { N } \bS \bS^\top - \frac{ \rho } { \barN } \barS \barS^\top.
\end{equation}

\par

\begin{algorithm}[t]
    \caption{Discriminative Feature-oriented Dictionary Learning}\label{alg:DFDL}
    \begin{algorithmic}
    \Function {$\bD^*$ = DFDL}{$\bY, \bar{\bY}, k, \rho$}
    \State \tb{INPUT:} {$\bY, \bar{\bY}$: collection of all in-class samples and complementary samples. $k$: number of bases in the dictionary. $\rho$: the regularization parameter.}
    \State 1. Choose initial $\bD^*$ and $L$ as in (\ref{eqn:findD0}) and (\ref{eqn:findL}).
    \While{not converged}
        \State 2. Fix $\bD = \bD^*$ and update $\bS, \bar{\bS}$ by solving (\ref{eqn:findS});
        \State 3. Fix $\bS, \bar{\bS}$, calculate:
        \begin{equation*}
            \bE = \frac{ 1 } { N } \boldsymbol{Y} \bS^\top - \frac{ \rho } { \bar{N} } \barX\barS^\top; \quad
             \Fb = \frac{ 1  } { N } \bS \bS^\top - \frac{ \rho } { \bar{N} } \barS \barS^\top.
        \end{equation*}
         \State 4. {Update $\bD$ from:
         \begin{equation*}
           \bD^*= \arg \min_{\bD} \Big\{-2\trace(\bE \bD^\top) + \trace\Big(\bD \big(\Fb - \lambda_{\min}(\Fb) \mathbf{I}\big) \bD^\top\Big)  \Big\}
         \end{equation*}
         \begin{equation*}
           \text{subject to:}  \|\bd_i\|_2^2 = 1, i = 1, 2, \dots, k.
         \end{equation*}
           }
      \EndWhile
      \State \tb{RETURN:} $\bD^*$
    \EndFunction
    \end{algorithmic}
\end{algorithm}

{The objective function in (\ref{subeqn:findD2}) is very similar to the objective function in the dictionary update stage problem in \cite{mairal2010online} except that it is not guaranteed to be convex. It is convex if and only if $\Fb$ is positive semidefinite. For the discriminative dictionary learning problem, the symmetric matrix $\Fb$ is {\em not} guaranteed to be positive semidefinite, even all of its eigenvalues are real. In the worst case, where $\Fb$ is negative semidefinite, the objective function in (\ref{subeqn:findD2}) becomes concave; if we apply the same dictionary update algorithm as in \cite{mairal2010online}, we will obtain its maximum solution instead of the minimum.}
	\par
{To deal with this situation, we propose a technique which convexifies the objective function based on the following observation.
}
	\par {If we look back to the main optimization problem stated in (\ref{eqn:findDopt}):
		\begin{equation*} 
		    \bD^* = \arg\min_{\bD} \left(\frac{1}{N} \min_{\|\bS\|_0 \leq L} \|\bY - \bD\bS\|_F^2 - \frac{\rho}{\bar{N}} \min_{\|\bar{\bS}\|_0 \leq L} \|\bar{\bY} - \bD \bar{\bS}\|_F^2\right),
		\end{equation*}
		we can see that if $\bD = \bmt \bd_1 & \bd_2 & \dots & \bd_k \emt $ is an optimal solution, then $\displaystyle \bD = \bmt \frac{\bd_1}{a_1} & \frac{\bd_2}{a_2} & \dots & \frac{\bd_k}{a_k} \emt$ is also an optimal solution as we multiply $j$-th rows of optimal $\bS$ and $ \bar{\bS}$ by $a_j$, where $a_j, j = 1, 2, \dots, k,$ are arbitrary nonzero scalars. Consequently, we can introduce constraints: $\|\bd_i\|_2^2 = 1, j = 1, 2, \dots, k$, without affecting optimal value of (\ref{subeqn:findD2}). With these constraints, $\trace(\bD \lambda_{\min} (\Fb)\mathbf{I}_k \bD^{\top}) = \lambda_{\min}(\Fb)\trace(\bD^{\top}\bD) = \lambda_{\min}(\Fb)\sum_{i = 1}^k \bd_i^{\top} \bd_i = k\lambda_{\min}(\Fb)$, where $\lambda_{\min}(\Fb)$ is the minimum eigenvalue of $\Fb$ and $\mathbf{I}_k$ denotes the identity matrix, is a constant. Substracting this constant from the objective function will not change the optimal solution to (\ref{subeqn:findD2}).}
{Essentially, the following problem in (\ref{eq:newProb}) is equivalent to (\ref{subeqn:findD2}):
\begin{equation} 
  \bD^* = \arg\min_{\bD}\{-2\trace(\bE\bD^{\top}) +  \trace\big(\bD(\Fb - \lambda_{\min}(\Fb)\mathbf{I}_k)\bD^{\top}\big)\}
  \label{eq:newProb}
\end{equation}
  $$\text{subject to:}  \|\bd_i\|_2^2 = 1, i = 1, 2, \dots, k.$$
The matrix $\hat{\Fb} = \Fb - \lambda_{\min}(\Fb)\mathbf{I}_k$ is guaranteed to be positive semidefinite since all of its eignenvalues now are nonnegative, and hence the objective function in (\ref{eq:newProb}) is convex. Now, this optimization problem is very similar to the dictionary update problem in \cite{mairal2010online}. Then, $\bD^*$ could be updated by the following iterations until convergence:
\begin{eqnarray}
    \label{eqn: updateuj}
    \bu_j &\leftarrow &\frac{1}{\hat{\Fb}_{j,j}}(\be_j - \bD\hat{\fb}_j) + \bd_j. \\
    \bd_j & \leftarrow & \frac{\bu_j}{\norm{\bu_j}_2}.
    \label{eqn: updatedj}
  \end{eqnarray}
where $\hat{\Fb}_{j,j}$ is the value of $\hat{\Fb}$ at coordinate $(j,j)$ and $\hat{\fb}_j$ denotes the $j$-th column of $\hat{\Fb}$.
}
\par
Our DFDL algorithm is summarized in Algorithm \ref{alg:DFDL}.
\subsection{Overall classification procedures for three datasets} 
\label{sec:overallclassification}
In this section, we propose a DFDL-based procedure for classifying images in three datasets.

\subsubsection{IBL and ADL datasets} 
\label{sub:ibladlprocedure}
The key idea in this procedure is that a healthy tissue image largely consists of healthy patches which cover a dominant portion of the tissue. This procedure is shown in Fig. \ref{fig: ibladlprocedure} and consists of the following three steps:
\par \noindent \textbf{Step 1}: \textit{Training DFDL bases for each class}. From labeled training images, training patches are randomly extracted (they might be overlapping). The size of these patches is picked based on pathologist input and/or chosen by cross validation\cite{Kohavi95astudy}. After we have a set of \textit{healthy} patches and a set of \textit{diseased} patches for training, class-specific DFDL dictionaries and the associated classifier are trained by using Algorithm \ref{alg:DFDL}.
\par\noindent \textbf{Step 2:}\label{} \textit{Learning a threshold $\theta$ for proportion of \textit{healthy} patches in one \textit{healthy} image}. Labeled training images are now divided into non-overlapping patches. Each of these patches is then classified using the DFDL classifier as described in Eq. (\ref{eqn:class1}) and (\ref{eqn:class2}). The main purpose of this step is to find the threshold $\theta$ such that healthy images have proportion of \textit{healthy} patches greater or equal to $\theta$ and diseased ones have proportion of \textit{diseased} patches less than $\theta$. {We can consider the proportion of healthy patches in one training image as its one-dimension feature. This feature is then put into a simple SVM to learn the threshold $\theta.$}


\par \noindent \textbf{Step 3:} \textit{Classifying test images}. For an unseen test image, we calculate the proportion $\tau$ of \textit{healthy} patches in the same way described in Step 2. Now, the identity of the image is determined by comparing the proportion $\tau$ to $\theta$. It is categorized as healthy (diseased) if $\tau \geq (<) \theta$.
The procedure readily generalizes to multi-class problems.
\label{sec:implementation_pipeline_for_}
\subsubsection{MVP detection problem in TCGA dataset} 
\label{sec:tcgaprocedure}
\begin{figure}
	\centering
  \includegraphics[width = 0.45\textwidth]{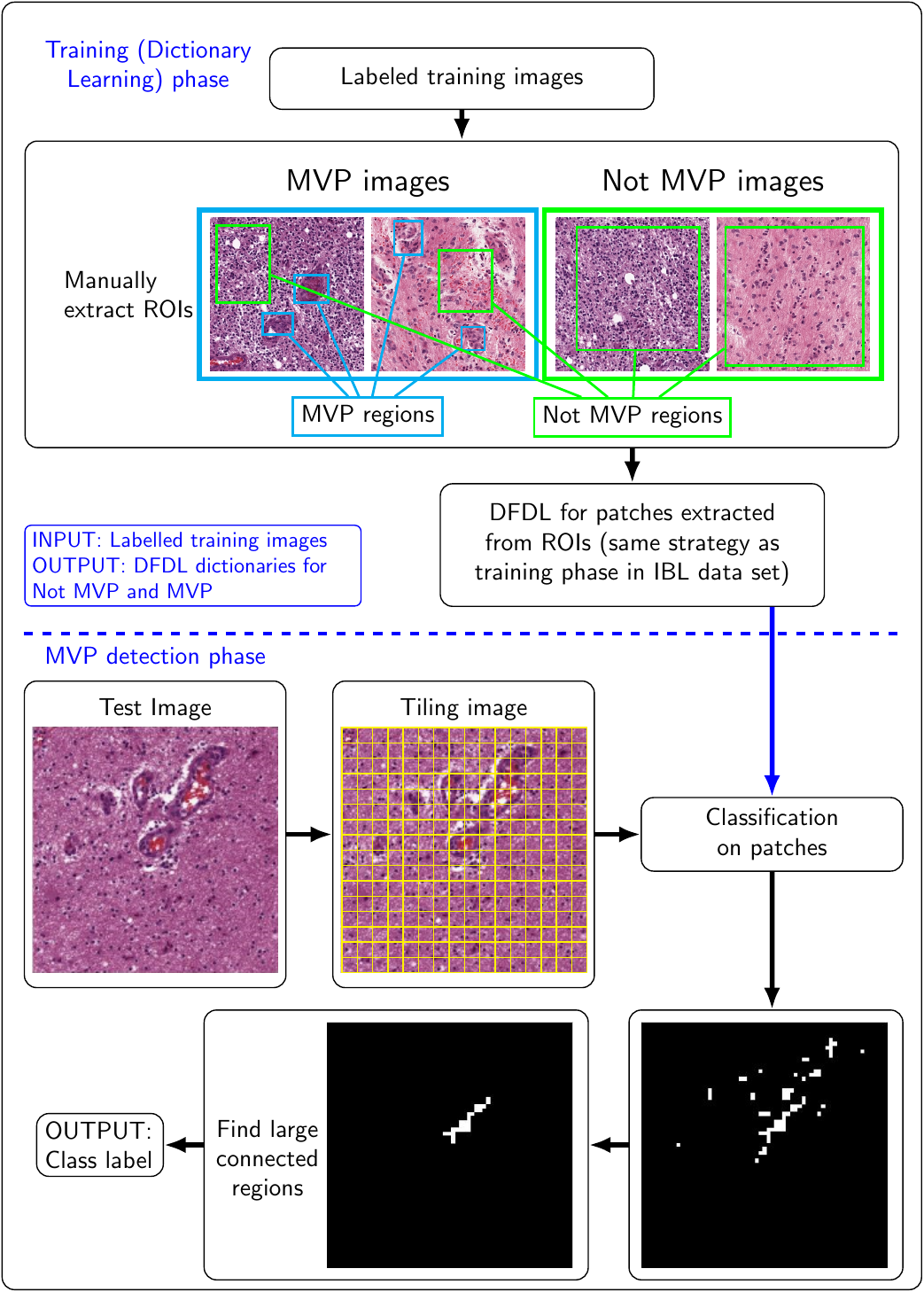}
  \caption{\small  MVP detection procedure}
  \label{fig:mvpdetection}
\end{figure}
\par As described earlier, MicroVascular Proliferation (MVP) is the presence of blood vessels in a tissue and it is an important indicator of a high-grade tumor in brain glioma. Essentially presence of one such region in the tissue image indicates the high-grade tumor. Detection of such regions in TCGA dataset is an inherently hard problem  and unlike classifying images in IBL and ADL datasets which are distinguishable by researching small regions, it requires more {effort} and investigation on larger connected regions. This is due to the fact that an MVP region may significantly vary in size and is usually surrounded by tumor cells which are actually benign or low grade. In addition, an MVP region is characterized by the presence of enlarged vessels in the tissue with different color shading and thick layers of cell rings inside the vessel (see Fig. \ref{fig:tcgasamples}).
We define a patch as \emph{MVP} if it lies entirely within an MVP region and as \emph{Not MVP} otherwise. We also define a region as Not MVP if it does not contain any MVP patch.
The procedure consists of two steps:
  \par \noindent\tb{Step 1:} \emph{Training phase}. From training data, MVP regions and Not MVP regions are manually extracted. Note that while MVP regions come from MVP images only, Not MVP regions might appear in all images. From these extracted regions, DFDL dictionaries are obtained in the same way as in step 1 of IBL/ADL classification procedure described in section \ref{sub:ibladlprocedure} and Fig. \ref{fig: ibladlprocedure}.
  \par{\noindent \tb{Step 2: } \emph{MVP detection phase:} A new unknown image is decomposed into non-overlapping patches. These patches are then classified using DFDL model learned before. After this step, we have a collection of patches classified as MVP. A region with large number of connected classified-as-MVP patches could be considered as an MVP region. If the final image does not contain any MVP region, we categorize the image as a Not MVP; otherwise, it is classified as MVP. {The definition of connected regions contains a parameter $m$, which is the number of connected patches. Depending on $m$, positive patches might or might not appear in the final step. Specifically, if $m$ is small, false positives tend to be determined as MVP patches; if $m$ is large, true positives are highly likely eliminated. To determine $m$, we vary it from 1 to 20 and compute its ROC curve for training images and then simply pick the point which is closest to the origin and find the {\em optimal} $m$. }}\label{par:choosem}
This procedure is visualized in Fig. \ref{fig:mvpdetection}. 


\par


\section{Validation and Experimental Results}
\label{sec:results} 
\begin{figure*}[t]
\centering
\includegraphics[width = 0.98\textwidth]{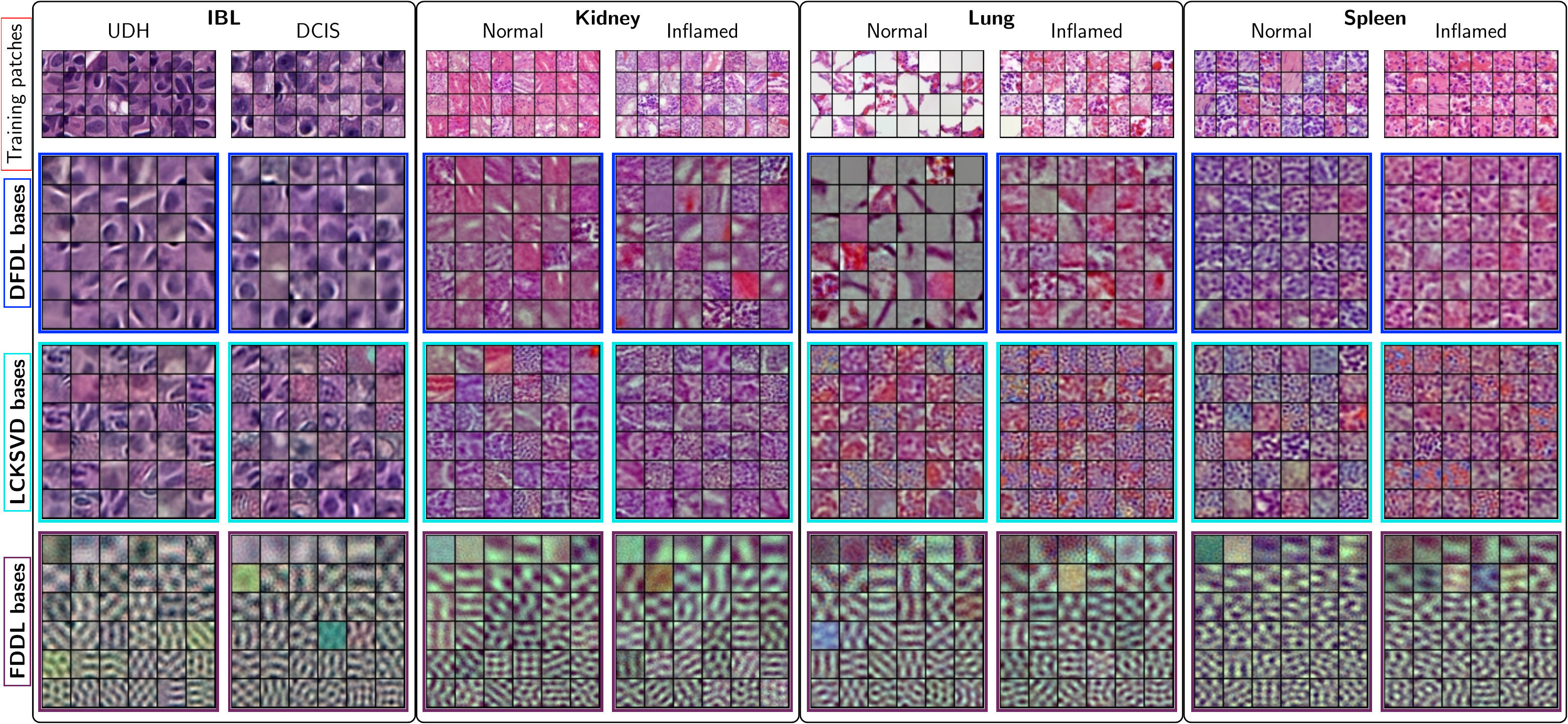}
\caption{\small  Example bases learned from different methods on different datasets. DFDL, LC-KSVD\cite{Zhuolin2013LCKSVD}, FDDL\cite{Meng2011FDDL} in IBL and ADL datasets. }
\label{fig:dictIKL}
\end{figure*}
\label{sec:experiment_results}
In this section, we present the experimental results of applying DFDL to three diverse histopathological image datasets and compare our results with different competing methods:
  \par$\bullet$ WND-CHARM\cite{Orlov2008,Shamir2008} in conjunction with SVM: this method combines state-of-the-art feature extraction and classification methods. We use the collection of features from WND-CHARM, which is known to be a powerful toolkit of features for medical images. While the original paper used weighted nearest neighbor as a classifier, we use a more powerful classifier (SVM \cite{CC01a}) to further enhance classification accuracy. We pick the most relevant features for histopathology\cite{Gurcan2009}, including but not limited to (color channel-wise) histogram information, image statistics, morphological features and wavelet coefficients from each color channel. The source code for WND-CHARM is made available by the National Institute of Health online at \url{http://ome.grc.nia.nih.gov/}.
  \par$\bullet$ SRC\cite{Wright2009SRC}: We apply SRC on the vectorization of the luminance channel of the histopathological images, as proposed initially for face recognition and applied widely thereafter.
  \par$\bullet$ SHIRC\cite{Srinivas2014SHIRC}: Srinivas \etal\cite{Srinivas2013,Srinivas2014SHIRC} presented a simultaneous sparsity model for multi-channel histopathology image representation and classification which extends the standard SRC\cite{Wright2009SRC} approach by designing three color dictionaries corresponding to the RGB channels. The MATLAB code for the algorithms is posted online at: \url{http://signal.ee.psu.edu/histimg.html}.
  \par$\bullet$ LC-KSVD\cite{Zhuolin2013LCKSVD} and FDDL\cite{Meng2011FDDL}: These are two well-known dictionary learning methods which were applied to object recognition such as face, digit, gender, vehicle, animal, etc, but to our knowledge, have not been applied to histopathological image classification. To obtain a fair comparison, dictionaries are learned on the same training patches. Classification is then carried out using the learned dictionaries on non-overlapping patches in the same way described in Section \ref{sec:overallclassification}.

  \par$\bullet$ Nayak's: In recent relevant work, Nayak \etal\cite{Nandita2013} proposed a patch-based method to solve the problem of classification of tumor histopathology via sparse feature learning. The feature vectors are then fed into  SVM to find the class label of each patch. 


\par
\subsection{Experimental Set-Up: Image Datasets} 
\label{subsec:experimental_setup_image_data_sets}
\par
\textbf{IBL dataset:} Each image contains a number of regions of interest (RoIs), and we have chosen a total of 120 images (RoIs), consisting of a randomly selected set of 20 images for training and the remaining 100 RoIs for test. Images are downsampled for computational  purposes such that size of a cell is around 20-by-20 (pixels). Examples of images from this dataset are shown in Fig. \ref{fig:iblsamples}.
Experiments in section \ref{subsec:validation_of_central_idea_overall_classification_accuracy} below are conducted with 10 training images per class, 10000 patches of size 20-by-20 for training per class, $k=500$ bases for each dictionary, $\lambda = 0.1$ and $\rho = 0.001$. {These parameters are chosen using cross-validation \cite{Kohavi95astudy}.}


\par \textbf{ADL dataset:} This dataset contains bovine histopathology images from three sub-datasets of kidney, lung and  spleen. Each sub-dataset consists of images of size $4000\times 3000$ pixels from two classes: healthy and inflammatory. Each class has around 150 images from which 40 images are chosen for training, the remaining ones are used for testing. Number of training patches, bases, $\lambda$ and $\rho$ are the same as in the IBL dataset. The classification procedure for IBL and ADL datasets is described in Section \ref{sub:ibladlprocedure}.


\textbf{TCGA dataset:} We use a total of 190 images (RoIs) (resolution {$3000\times 3000$}) from the TCGA, in which 57 images contain MVP regions and 133 ones have no traces of MVP. From each class, 20 images are randomly selected for training. The classification procedure for this dataset is described in Section \ref{sec:tcgaprocedure}. 
\par {Each tissue specimen in these datasets is fixed on a scanning bed and digitized using a digitizer at 40$\times$ magnification.}
\subsection{Validation of Central Idea: Visualization of Discovered Features} 
\label{subsec:validation_of_central_idea_overall_classification_accuracy}
\begin{figure}[t]
\centering
  \includegraphics[width=0.48\textwidth]{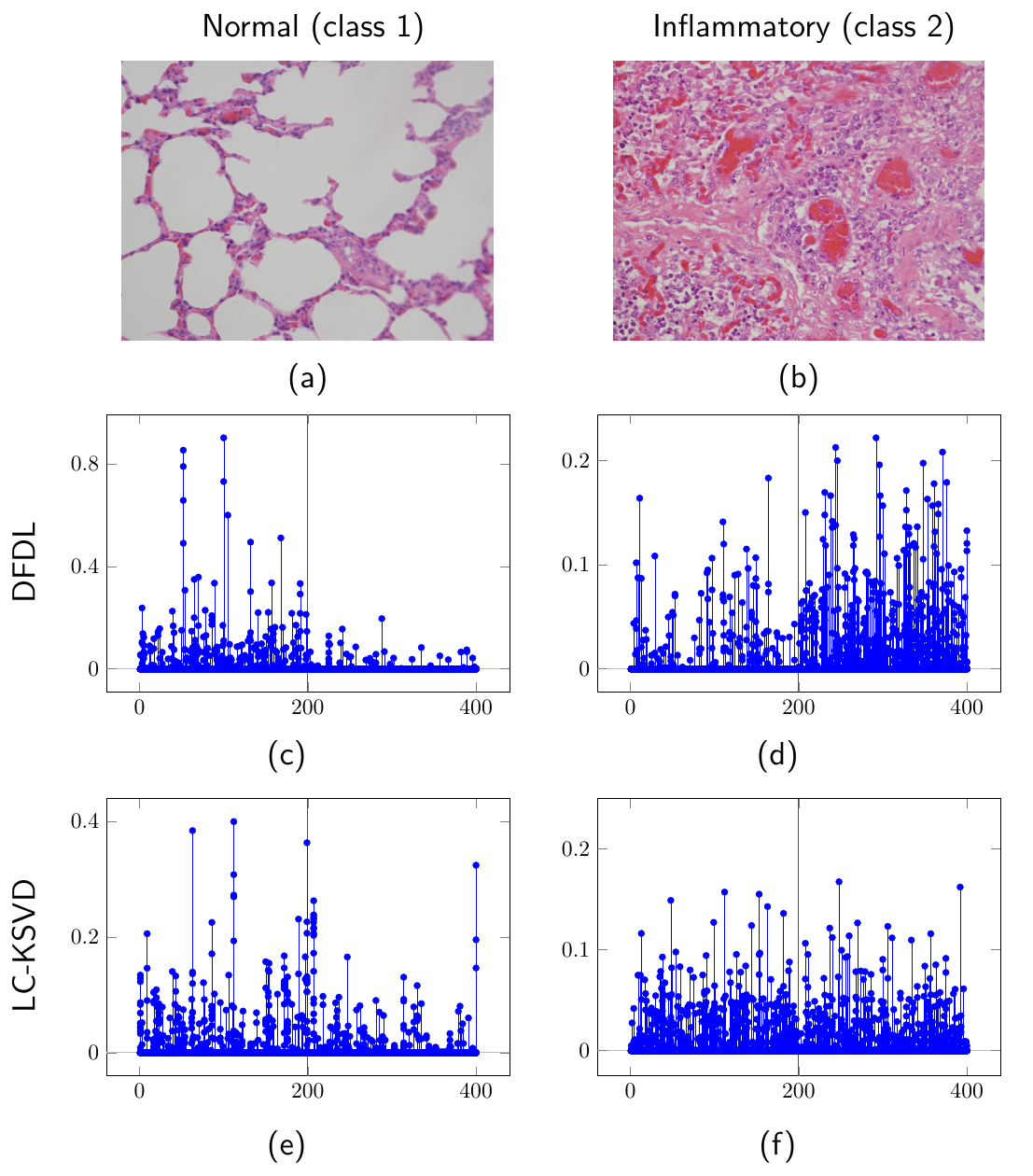}
 \caption{\small {Example of sparse codes using DFDL and LC-KSVD approaches on lung dataset. Left: normal lung (class 1). Right: inflammatory lung (class 2). Row 1: test images. Row 2: Sparse codes visualization using DFDL. Row 3: Sparse codes visualization using LC-KSVD. $x$ axis indicates the dimensions of sparse codes with codes on the left of red lines corresponding to bases of class 1, those on the right are in class 2. $y$ axis demonstrates values of those codes. In one vertical line, different dots represent values of non-zeros coefficients of different patches.}
 \vspace{-0.15in}}
  \label{fig: visual_coef }
\end{figure}

This section provides experimental validation of the central hypothesis of this paper: by imposing sparsity constraint on forcing intra-class differences to be small, while simultaneously emphasizing inter-class differences, the class-specific bases obtained are discriminative.


\par {Example bases obtained by different dictionary learning methods are visualized in Fig. \ref{fig:dictIKL}. By visualizing these bases, we emphasize that our DFDL is able to look for discriminative visual features from which pathologists could understand the reasons behind diseases. In the spleen dataset for example, it is really difficult to realize the differences between two classes by human eyes. However, by looking at DFDL learned bases, we can see that the distribution of cells in two classes are different such that a larger number of cells appears in a normal patch. These differences may provide pathologists one visual cue to classify these images without advanced tools. Moreover, for IBL dataset, UDH bases visualize elongated cells with sharp edges while DCIS bases present more rounded cells with blurry boundaries, which is consistent with their descriptions in \cite{Srinivas2014SHIRC} and \cite{Dundar2011}; for ADL-Lung, we observe that a healthy lung is characterized by large clear openings of the alveoli, while in the inflamed lung, the alveoli are filled with bluish-purple inflammatory cells. This distinction is very clear in the bases learned from DFDL where white regions appear more in normal bases than in inflammatory bases and no such information can be deduced from LC-KSVD or FDDL bases. In comparison, FDDL fails to discover discriminative visual features that are interpretable and LC-KSVD learns bases with the inter-class differences being less significant than DFDL bases. Furthermore, these LC-KSVD bases do not present key properties of each class, especially in lung dataset.} 
\par
{To understand more about the significance of discriminative bases for classification, let us first go back to SRC~\cite{Wright2009SRC}. For simplicity, let us consider a problem with two classes with corresponding dictionaries $\bD_1$ and $\bD_2$. The identity of a new patch $\by$, which, for instance, comes from class 1, is determined by equations (\ref{eqn:class1}) and (\ref{eqn:class2}). In order to obtain good results, we expect most of active coefficients to be present in $\delta_1(\hat{\bs})$. For $\delta_2(\hat{\bs})$, its non-zeros, if they exists should have small magnitude. Now, suppose that one basis, $\bd_1$, in $\bD_1$ looks very similar to another basis, $\bd_2$, in $\bD_2$. When doing sparse coding, if one patch in class 1 uses $\bd_1$ for reconstruction, it is highly likely that a similar patch $\by$ in the same class uses $\bd_2$ for reconstruction instead. This misusage may lead to the case $\|\by -\bD_1 \delta_1(\hat{\bs})\| > \|\by - \bD_2 \delta_2(\hat{\bs})\|$, resulting in a misclassified patch. For this reason, the more discriminative bases are, the better the performance. }

\par {To formally verify this argument, we do one experiment on one normal and one inflammatory image from lung dataset in which the differences of DFDL bases and LCKSVD bases are most significant. From these images, patches are extracted, then their sparse codes are calculated using two dictionaries formed by DFDL bases and LC-KSVD bases. Fig. \ref{fig: visual_coef } demonstrates our results. Note that the plots in Figs.\ \ref{fig: visual_coef }c) and d) are corresponding to DFDL while those in Figs. \ref{fig: visual_coef }e) and f) are for LC-KSVD. Most of active coefficients in Fig. \ref{fig: visual_coef }c) are gathered on the left of the red line, and their values are also greater than values on the right. This means that $\bD_1$ contributes more to reconstructing the lung-normal image in Fig. \ref{fig: visual_coef }a) than $\bD_2$ does. Similarly, most of active coefficients in Fig. \ref{fig: visual_coef }d) locate on the right of the vertical line. This agrees with what we expect since the image in Fig. \ref{fig: visual_coef }a) belongs to class 1 and the one in Fig. \ref{fig: visual_coef }b) belongs to class 2. On the contrary, for LC-KSVD, active coefficients in Fig. \ref{fig: visual_coef }f) are more uniformly distributed on both sides of the red line, which adversely affects classification. In Fig. \ref{fig: visual_coef }e), although active coefficients are strongly concentrated to the left of the red line, this effect is even more pronounced with DFDL, i.e.\ in  Fig. \ref{fig: visual_coef }c).
}

\begin{figure}[t]
\centering
  \includegraphics[width=0.45\textwidth]{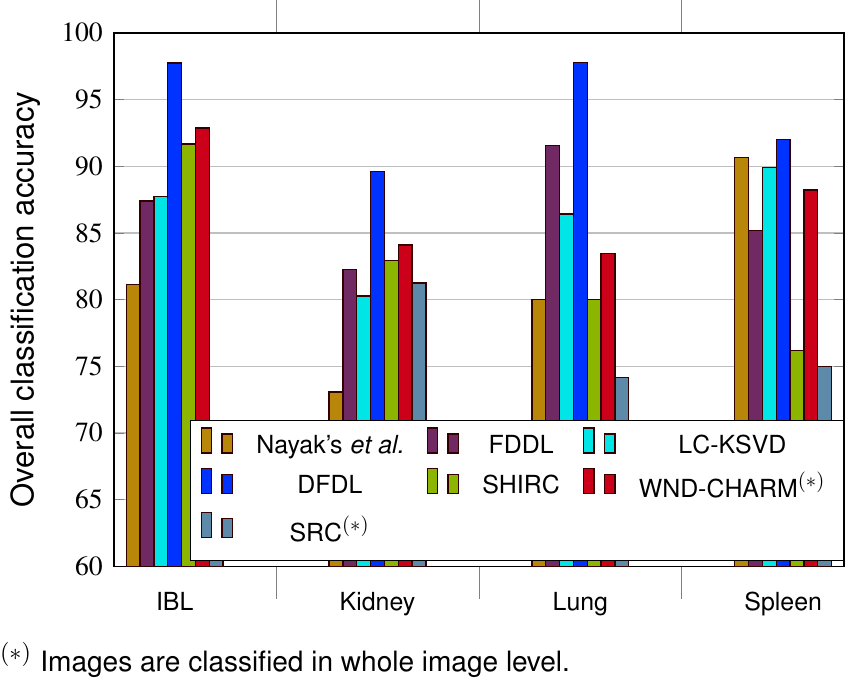}
 \caption{\small Bar graphs indicating the overall classification accuracies ($\%$) of the competing methods.}
\vspace{-0.15in}
  \label{fig: OverallAcc}
\end{figure}

\begin{figure}[t]
\centering
  \includegraphics[width=0.48\textwidth]{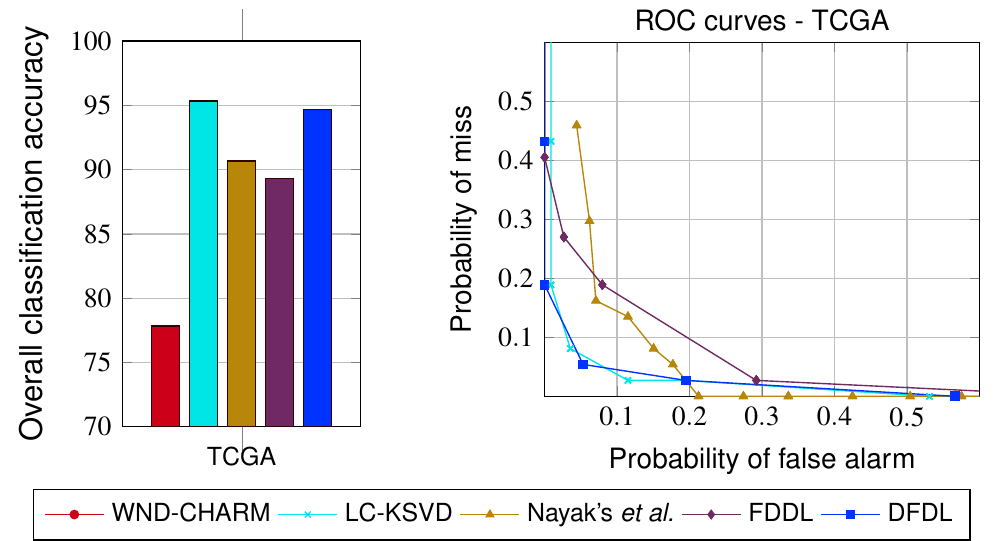}
 \caption{\small Bar graphs (left) indicating the overall classification accuracies ($\%$) and the receiver operating characteristic (right) of the competing methods for TCGA dataset.}
\vspace{-0.15in}
  \label{fig: OverallAccTCGA}
\end{figure}
\par
\begin{table}[t]
\centering
\caption{ CONFUSION MATRIX: IBL. }
\label{tab: IBL_confusion}
\begin{tabular}{|c|c|c||l|}
\hline
Class                 & UDH            & DCIS           & Method  \\ \hline
\multirow{8}{*}{UDH}  & 91.75          & 8.25          & WND-CHARM$^{(*)}$ \cite{Shamir2008}   \\
                      & 68.00          & 32.00          & SRC$^{(*)}$     \cite{Wright2009SRC}     \\
                      & \textit{\textbf{93.33}}   & 6.67           & SHIRC    \cite{Srinivas2014SHIRC} \\
                      & 84.80          & 15.20          & FDDL    \cite{Meng2011FDDL}    \\
                      & 90.29          & 9.71           & LC-KSVD \cite{Zhuolin2013LCKSVD}     \\
                      & 85.71          & 14.29          & Nayak's \etal\cite{Nandita2013}   \\
                      & \textbf{96.00} & 4.00           & DFDL         \\ \hline
\multirow{8}{*}{DCIS} & 5.77           & \textit{\textbf{94.23}}          & WND-CHARM$^{(*)}$ \cite{Shamir2008}    \\
                      & 44.00          & 56.00          & SRC$^{(*)}$       \cite{Wright2009SRC}     \\
                      & 10.00          & 90.00          & SHIRC   \cite{Srinivas2014SHIRC} \\
                      & 10.00          & 90.00          & FDDL    \cite{Meng2011FDDL}        \\
                      & 14.86          & 85.14          & LC-KSVD   \cite{Zhuolin2013LCKSVD}      \\
                      & 23.43          & 76.57          & Nayak's \etal\cite{Nandita2013}      \\
                      & 0.50           & \textbf{99.50} & DFDL         \\ \hline
\end{tabular}
\begin{center}
	\begin{tablenotes}
	      {\small      $^{(*)}$ Images are classified in whole image level.}
  	\end{tablenotes}
\end{center}
\end{table}
\subsection{Overall Classification Accuracy} 
\label{sub:overall_classification_accuracy}

 To verify the performance of our idea, for IBL and ADL datasets, we present overall classification accuracies in the form of bar graphs in Fig. \ref{fig: OverallAcc}. It is evident that DFDL outperforms other methods in both  datasets. Specifically, in IBL and ADL Lung, the overall classification accuracies of DFDL are over 97.75$\%$, the next best rates come from WND-CHARM (92.85$\%$ in IBL) and FDDL (91.56$\%$ in ADL-Lung), respectively, and much higher than those reported in \cite{Srinivas2014SHIRC} and our own previous results in \cite{vu2015dfdl}. In addition, for ADL-Kidney and ADL-Spleen, our DFDL also provides the best result with accuracy rates being nearly 90$\%$ and over 92$\%$, respectively.


 For the TCGA dataset, overall accuracy of competing methods are shown in Fig. \ref{fig: OverallAccTCGA}, which reveals that DFDL performance is the second best, bettered only by LC-KSVD and by less than 0.67$\%$ (i.e.\ one more misclassified image for DFDL). 

\subsection{Complexity analysis} 
\label{sec:complexity_analysis}

In this section, we compare the computational complexity for the proposed DFDL and competing dictionary learning methods: LC-KSVD\cite{Zhuolin2013LCKSVD}, FDDL\cite{Meng2011FDDL}, and Nayak's\cite{Nandita2013}. The complexity for each dictionary learning method is estimated as the (approximate) number of operations required by each method in learning the dictionary (see Appendix for details).
From Table \ref{tab:complexity}, it is clear that the proposed DFDL is the least expensive computationally. Note further, that the final column of Table \ref{tab:complexity} shows {\em actual run times} of each of the methods. The parameters were as follows: $c = 2$ (classes), $k = 500$ (bases per class), $N = 10,000$ (training patches per class), data dimension $d = 1200$ (3 channels $\times20\times20$), sparsity level $L = 30$. The run time numbers in the final column of Table \ref{tab:complexity} are in fact consistent with numbers provided in Table \ref{tab:operations}, which are calculated by plugging the above parameters into the second column of Table \ref{tab:complexity}.

\begin{table}[t]
\centering
\caption{ {Complexity analysis for different dictionary learning methods.}}
\label{tab:complexity}
  \begin{tabular}{|l|l|l|}
  \hline
  Method & Complexity & Running time \\ \hline
  DFDL &$c^2kN(2d + L^2)$  & $\sim$ 0.5 hours\\ \hline
  LC-KSVD\cite{Zhuolin2013LCKSVD} & $c^2kN(2d + 2ck + L^2)$& $\sim$ 3 hours \\ \hline
  Nayak's \etal\cite{Nandita2013}$^{(*)}$ &$c^2kN(2d + 2qck) + c^2dk^2$& $\sim$ 8 hours \\ \hline
  FDDL\cite{Meng2011FDDL}$^{(*)}$ & $c^2kN(2d + 2qck) + c^3dk^2$& > 40 hours\\ \hline
  \end{tabular}
  \begin{tablenotes}
      \small
      \item $^{(*)}$$q$ is the number of iterations required for $l_1$-minimization in sparse coding step.
    \end{tablenotes}
\end{table}
\begin{table}[t]
\centering
\caption{Estimated number of operations required in different dictionary learning methods.}
\label{tab:operations}
  \begin{tabular}{|l|l|l|l|}
  \hline
  Method &  $q = 1$ &  $q = 3$        &              $q = 10$\\ \hline
  DFDL      & $6.6 \times 10^{10}$    & $6.6 \times 10^{10}$    & $6.6 \times 10^{10}$    \\ \hline
  LC-KSVD\cite{Zhuolin2013LCKSVD}  & $1.06 \times 10^{11}$   & $1.06 \times 10^{11}$   & $1.06 \times 10^{11}$   \\ \hline
  Nayak's \etal\cite{Nandita2013}     & $8.92\times 10^{10}$    & $1.692 \times 10^{11}$  & $4.492 \times 10^{11}$\\ \hline
  FDDL\cite{Meng2011FDDL}      & $9.04 \times 10^{10}$   & $1.704 \times 10^{11}$  & $4.504 \times 10^{11}$\\ \hline
  \end{tabular}
\end{table}


\begin{table*}[ht!]

\centering
\caption{ CONFUSION MATRIX: ADL ($\%$).}
\label{tab: adl_confusion}
\begin{tabular}{|c||c|c||c|c||c|c||l|}
\hline
             & \multicolumn{2}{|c||}{Kidney}                       & \multicolumn{2}{|c||}{Lung}                        & \multicolumn{2}{|c||}{Spleen}                      &           \\ \hline
Class        & Health                 & inflammatory            & Health                 & inflammatory            & Health                 & inflammatory            & \multicolumn{1}{c|}{Method}    \\ \hline \hline
\multirow{8}{*}{Health}
             & 83.27                   & 16.73                   & 83.20                   & 16.80                   & 87.23                   & 12.77                   & WND-CHARM$^{(*)}$ \cite{Shamir2008}   \\
             & \textit{\textbf{87.50}} & 12.50                   & 72.50                   & 27.50                   & 70.83                   & 29.17                   & SRC$^{(*)}$    \cite{Wright2009SRC}     \\
             & 82.50                   & 17.50                   & 75.00                   & 25.00                   & 65.00                   & 35.00                   & SHIRC    \cite{Srinivas2014SHIRC} \\
             & 83.26                   & 16.74                   & \textit{\textbf{93.15}}                   & 6.85                    & 86.94                   & 13.06                   & FDDL    \cite{Meng2011FDDL}    \\
             & 86.84                   & 13.16                   & 85.59                   & 15.41                   & \textit{\textbf{89.75}}                   & 10.25                   & LC-KSVD \cite{Zhuolin2013LCKSVD}     \\
             & 73.08                   & 26.92                   & 89.55                   & 10.45                   & 86.44                   & 13.56                   & Nayak's \etal \cite{Nandita2013}  \\
             & \textbf{88.21}          & 11.79                   & \textbf{96.52}          & 3.48                    & \textbf{92.88}          & 7.12                    & DFDL      \\ \hline
\multirow{8}{*}{inflammatory}
             & 14.22                   & 85.78                   & 14.31                   & 83.69                   & 10.48                   & 89.52                   & WND-CHARM$^{(*)}$ \cite{Shamir2008}   \\
             & 25.00                   & 75.00                   & 24.17                   & 75.83                   & 20.83                   & 79.17                   & SRC$^{(*)}$    \cite{Wright2009SRC}     \\
             & 16.67                   & \textit{\textbf{83.33}}                   & 15.00                   & 85.00                   & 11.67                   & 88.33                   & SHIRC    \cite{Srinivas2014SHIRC} \\
             & 19.88                   & 80.12                   & 10.00                   & \textit{\textbf{90.00}}                   & 8.57                    & 91.43                        & FDDL    \cite{Meng2011FDDL}    \\
             & 19.25                   & 81.75                   & 10.89                   & 89.11                   & 8.57                    & 91.43                   & LC-KSVD \cite{Zhuolin2013LCKSVD}     \\
             & 26.92                   & 73.08                   & 25.90                   & 74.10                   & 6.05                    & \textbf{93.95}          & Nayak's \etal\cite{Nandita2013}   \\
             & 9.92                    &\textbf{90.02} & 2.57                    & \textbf{97.43}          & 7.89                    & \textit{\textbf{92.01}} & DFDL      \\ \hline
\end{tabular}
\begin{center}
\begin{tablenotes}
\centering
      {\small      $^{(*)}$ Images are classified in whole image level.}
      \end{tablenotes}
\end{center}
\end{table*}

\subsection{Statistical Results: Confusion Matrices and ROC Curves} 
\label{subsec:detailed_results_confusion_matrices_and_roc_curves}



\begin{table}[t]
\centering
\caption{ CONFUSION MATRIX: TCGA ($\%$).}
\vspace{-0.1in}
\label{tab: tcga_confution}
\begin{tabular}{|c|c|c||l|}
\hline
     Class               & Not MVP & MVP   & Method    \\ \hline
\multirow{4}{*}{Not VMP} & 76.68   & 23.32 & WND-CHARM\cite{Shamir2008} \\
                         & 92.92   & 7.08  & Nayak's \etal\cite{Nandita2013}   \\
                         & \textbf{96.46}   & 3.54  & LC-KSVD\cite{Zhuolin2013LCKSVD}   \\
                         & {92.04}   & 7.96  & FDDL\cite{Meng2011FDDL}   \\
                         & \textit{\textbf{94.69}}   & 5.31  & DFDL      \\ \hline
\multirow{4}{*}{MVP}     & 21.62   & 78.38 & WND-CHARM\cite{Shamir2008} \\
                         & 16.22   & 83.78 & Nayak's \etal\cite{Nandita2013}   \\
                         & 8.10    & \textit{\textbf{91.90}} & LC-KSVD\cite{Zhuolin2013LCKSVD}   \\
                         & 18.92    & {81.08} & FDDL\cite{Meng2011FDDL}   \\
                         & 5.41   & {\textbf{94.59}} & DFDL      \\ \hline
\end{tabular}
\end{table}

\begin{figure*}[t]
\centering
  \includegraphics[width=\textwidth]{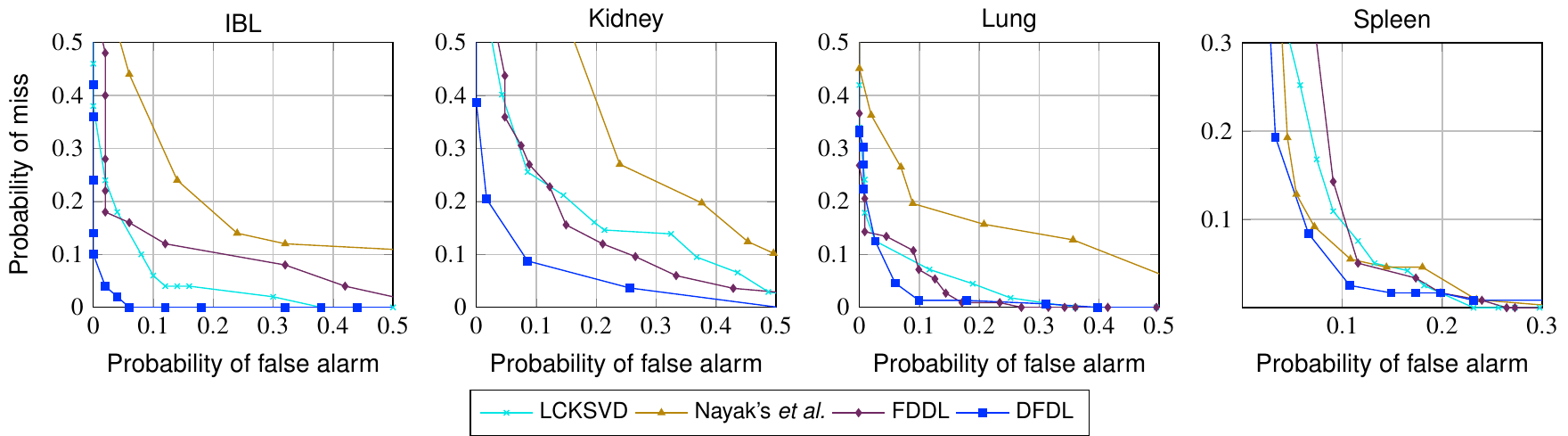}
 \caption{\small Receiver operating characteristic (ROC) curves for different organs, methods, and datasets (IBL and ADL).}
  \label{fig: ibladlroc}
\end{figure*}

\begin{figure*}[t]
\centering
  \includegraphics[width=\textwidth]{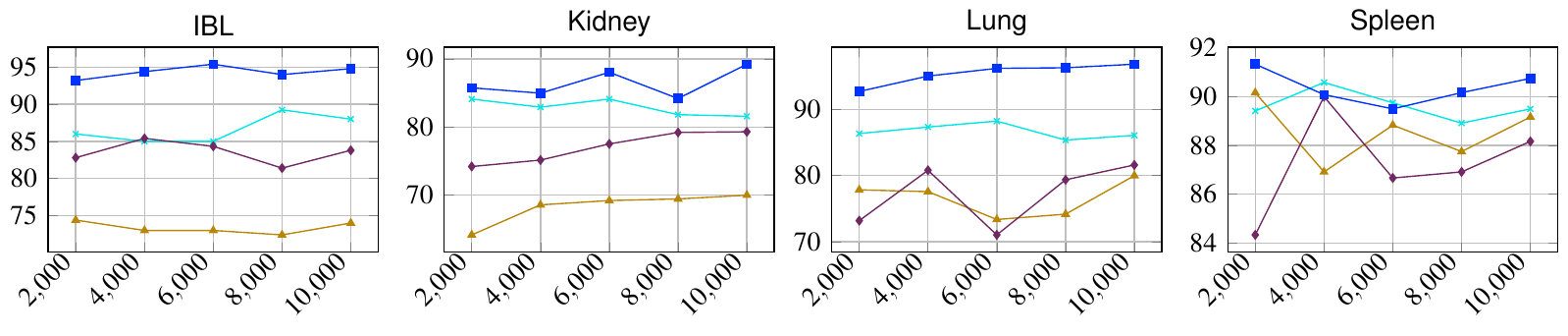}\\
  \includegraphics[width=\textwidth]{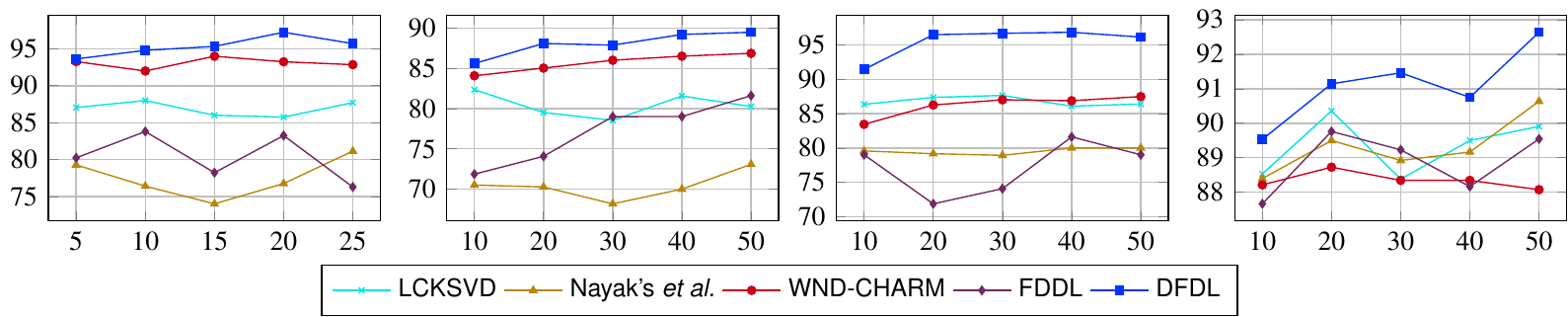}
 \caption{Overall classification accuracy ($\%$) as a function of training set size per class. Top row: number of training patches. Bottom row: number of training images.}
\vspace{-0.2in}
  \label{fig: CompareNimgs}
\end{figure*}
\begin{figure*}[t]
\centering
\includegraphics[width=\textwidth]{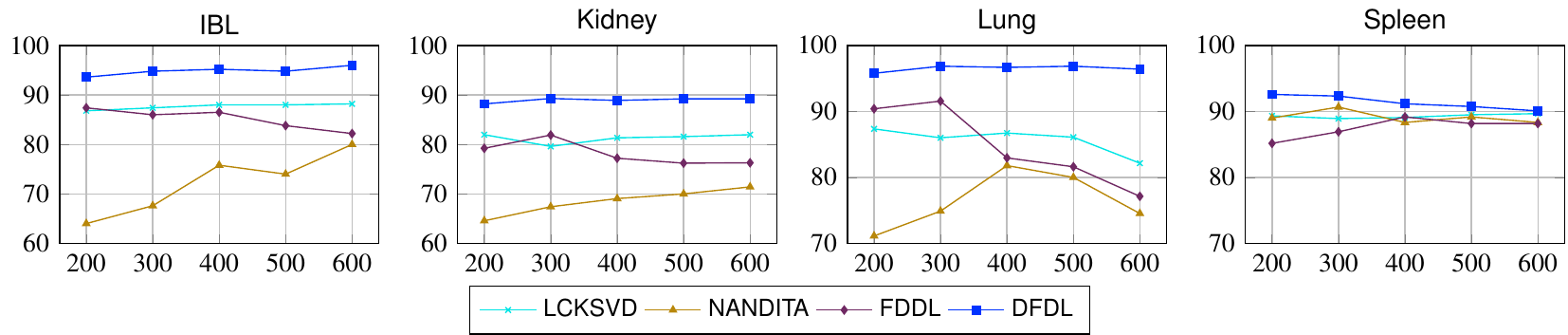}
 \caption{{Overall classification accuracy ($\%$) as a function of number of training bases.}}
  \label{fig: CompareNbases}
\end{figure*}

Next, we present a more elaborate interpretation of classification performance in the form of confusion matrices and ROC curves. Each row of a confusion matrix refers to the actual class identity of test images and each column indicates the classifier output. Table \ref{tab: IBL_confusion}, \ref{tab: adl_confusion} and \ref{tab: tcga_confution} show the mean confusion matrices for all of three dataset. In continuation of trends from Fig. \ref{fig: OverallAcc}, in Table \ref{tab: adl_confusion}, DFDL offers the best disease detection accuracy in almost all datasets for each organ, while maintaining high classification accuracy for healthy images.
\par
Typically in medical image classification problems, pathologists desire algorithms that reduce the probability of miss (diseased images are misclassified as healthy ones) while also ensuring that the false alarm rate remains low. However, there is a trade-off between these two quantities, conveniently described using receiver operating characteristic (ROC) curves. Fig. \ref{fig: ibladlroc} and Fig. \ref{fig: OverallAccTCGA} (right) show the ROC curves for all three datasets. The lowest curve (closest to the origin) has the best overall performance and the optimal operating point minimizes the sum of the miss and false alarm probabilities. It is evident that ROC curves for DFDL perform best in comparison to those of other state-of-the-art methods.
\par {\textbf{Remark:} Note for ROC comparisons, we compare the different flavors of dictionary learning methods (the proposed DFDL, LC-KSVD, FDDL and Nayak's), this is because as Table \ref{tab: tcga_confution} shows, they are the most competitive methods. Note for the IBL and ADL datasets, $\theta$, as defined in Fig. \ref{fig: ibladlprocedure}, is changed from 0 to 1 to acquire the curves; whereas for the TCGA dataset, number of connected classified-as-MVP patches, $m$, is changed from 1 to 20 to obtain the curves.} It is worth re-emphasizing that DFDL achieves these results even as its complexity is lower than competing methods.

\subsection{Performance vs. size of training set} 
\label{sec:performance_vs_size_of_training_set}
Real-world histopathological classification tasks must often contend with lack of availability of large training sets. To understand training dependence of the various techniques, we present a comparison of overall classification accuracy as a function of the training set size for the different methods. We also present a comparison of classification rates as a function of the number of training patches for different dictionary learning methods\footnote{Since WND-CHARM is applied in the whole image level, there is no result for it in comparison of training patches.}. In Fig. \ref{fig: CompareNimgs}, overall classification accuracy is reported for IBL and ADL datasets corresponding to five scenarios. It is readily apparent that DFDL exhibits the most graceful decline as training is reduced.
\subsection{{Performance vs. number of training bases}} 
\label{sub:performance_vs_number_of_training_bases}

{We now compare the behavior of each dictionary learning method as the number of bases in each dictionary varies from 200 to 600 (with patch size being fixed at $20\times20$ pixels). Results reported in Fig. \ref{fig: CompareNbases} confirm that DFDL again outperforms other methods. In general, overall accuracies of DFDL on different datasets remain high when we reduce number of training bases. Interpreted another way, these results illustrate that DFDL is fairly robust to changes in parameters, which is a highly desirable trait in practice.
}
\section{{Discussion and Conclusion}} 
\label{sec:Conclusion}
\par
{In this paper, we address the histopathological image classification problem from a feature discovery and dictionary learning standpoint. This is a very important and challenging problem and the main challenge comes from the geometrical richness of tissue images, resulting in the difficulty of obtaining reliable discriminative features for classification. Therefore, developing a framework capable of capturing this structural richness and  being able to discriminate between different types is investigated and to this end, we propose the DFDL method which learns discriminative features for histopathology images. Our work aims to produce a more versatile histopathological image classification system through the design of discriminative, class-specific dictionaries which is hence capable of automatic feature discovery using example training image samples.
}\par
{Our DFDL algorithm learns these dictionaries by leveraging the idea of sparse representation of in-class and out-of-class samples. This idea leads to an optimization problem which encourages intra-class similarities and emphasizes the inter-class differences. Ultimately, the optimization in \eqref{subeqn:findD2} is done by solving the proposed equivalent optimization problem using a convexifying trick. Similar to other dictionary learning (machine learning approaches in general), DFDL also requires a set of regularization parameters. Our DFDL requires only one parameter, $\rho$, in its training process which is chosen by cross validation\cite{Kohavi95astudy} -- plugging different sets of parameters into the problem and selecting one which gives the best performance on the validation set. In the context of application of DFDL to real-world histopathological image slides, there are quite a few other settings should be carefully chosen, such as patch size, tiling method, number of connected components in the MVP detection etc. Of more importance is the patch size to be picked for each dataset which is mostly determined by consultation with the medical expert in the specific problem under investigation and the type of features that we should be looking for. For simplicity we employ regular tiling; however, using prior domain knowledge this may be improved. For instance in the context of MVP detection, informed selection of patch locations using existing disease detection and localization methods such as \cite{Mousavi2015JPI} can be used to further improve the detection of disease.
}\par
{Experiments are carried out on three diverse histopathological datasets to show the broad applicability of the proposed DFDL method. It is illustrated our method is competitive with or outperforms state of the art alternatives, particularly in the regime of realistic or limited training set size. It is also shown that with minimal parameter tuning and algorithmic changes, DFDL method can be easily applied on different problems with different natures which makes it a good candidate for automated medical diagnosis instead of using customized and problem specific frameworks for every single diagnosis task. We also make a software toolbox available to help deploy DFDL widely as a diagnostic tool in existing histopathological image analysis systems. Particular problems such as grading and detecting specific regions in histopathology may be investigated using our proposed techniques.
}\par

\newpage
\appendix{}
\label{sub:complexity_analysis}
\begin{center}
{	COMPLEXITY ANALYSIS}
\end{center}

{In this section, we compare the computational complexity for the proposed DFDL and competing dictionary learning methods: LC-KSVD\cite{Zhuolin2013LCKSVD}, FDDL\cite{Meng2011FDDL}, and Nayak's\cite{Nandita2013}. The complexity for each dictionary learning method is estimated as the (approximate) number of operations required by each method in learning the dictionary.
For simplicity, we assume that number of training samples, number of dictionary bases in each class are the same, which means: $N_i = N_j = N, k_i = k_j = k, \forall i,j = 1, 2, \dots, c$, and also $L_i = L_j = L, \forall i,j = 1, 2, \dots, c.$ For the consitence, we have changed notations in those methods by denoting $\bY$ as training samples and $\bS$ as the sparse code.
}
\par

{In most of dictionary learning methods, the complexity of sparse coding step, which is often a $l_0$ or $l_1$ minimization problem, dominates that of dictionary update step, which is typically solved by either block coordinate descent\cite{mairal2010online} or singular value decomposition\cite{Aharon2006KSVD}. Then, in order to compare the complexity of different dictionary learning methods, we focus on comparing the complexity of sparse coding steps in each iteration.} 
\subsection{{Complexity of the DFDL}} 
\label{sec:complexity_of_the_ompcite_tropp2007signal}
{The most expensive computation in DFDL is solving an Orthogonal Matching Pursuit (OMP \cite{tropp2007signal}) problem. Given a set of samples $\bY \in \R^{d \times N}$, a dictionary $\bD \in \R^{d\times k} $ and sparsity level $L$, the OMP problem is:
\begin{equation*}
    \bS^* = \arg\min_{\|\bS\|_0 \leq L} \|\bY - \bD\bS\|_F^2.
\end{equation*}
 R. Rubinstein \etal\cite{Rubinstein2008} reported the complexity of Batch-OMP when the dictionary is stored in memory in its entirety as:
     $T_{\text{b-omp}} = N(2dk + L^2k + 3Lk + L^3) + dk^2.$
Assuming an asymptotic behavior of $L \ll k \approx d \ll N $, the above expression can be simplified to:
\begin{equation}
    T_{\text{b-omp}} \approx N(2dk + L^2k) = kN(2d + L^2).
\end{equation}
 This result will also be utilized in analyzing complexity of LC-KSVD.}\par
\par
{The sparse coding step in our DFDL consists of solving $c$ sparse coding problems:
    $\hat{\bS} = \arg\min_{\|\bS\|_0 \leq L} \norm{\hat{\bY} - \bD_i\hat{\bS}_i}_F^2.$
With $\hat{\bY} \in \R^{d\times cN}, \bD_i \in \R^{d\times k}$, each problem has complexity of $k(cN)(2d + L^2)$. Then the total complexity of these $c$ problems is:
    $T_{\text{DFDL}} \approx c^2kN(2d + L^2)$.
}\subsection{{Complexity of LC-KSVD}} 
\label{sec:complexity_of_lc_ksvdcite_zhuolin2013lcksvd}
{We consider LC-KSVD1 only (LC-KSVD2 has a higher complexity) whose optimization problem is written as \cite{Zhuolin2013LCKSVD}:
\begin{equation*}
    (\bD, \bA, \bS) = \arg\min_{\bD, \bA, \bS}\norm{\bY -\bD\bS}_F^2 + \alpha\norm{\bQ - \bA\bS}_F^2 \text{~s.t.~} \norm{\bs_i}_0 \leq L.
\end{equation*}
and it is rewritten in the K-SVD form:
\begin{equation}
  \label{eqn:lcksvd}
    (\bD, \bA, \bS) = \arg\min_{\bD, \bA, \bS}\norm{\bmt \bY \\ \sqrt{\alpha}\bQ \emt  -\bmt \bD \\ \sqrt{\alpha}\bA \emt\bS}_F^2 \text{~s.t.~} \norm{\bs_i}_0 \leq L.
\end{equation}
Since $\bQ \in \R^{ck \times cN}$ and $\bA \in \R^{ck \times ck}$, $\tilde{\bY} = \bmt \bY \\ \sqrt{\alpha}\bQ \emt \in \R^{(d + ck) \times cN}$ and $\tilde{\bD} = \bmt \bD \\ \sqrt{\alpha}\bA \emt \in \R^{(d + ck)\times ck} $. Neglecting the computation of scalar multiplications, the complexity of (\ref{eqn:lcksvd}) is:
\begin{equation*}
    T_{\text{LC-KSVD}} \approx (ck)(cN)(2(d+ck)+L^2) = c^2kN(2d + 2ck + L^2).
\end{equation*}
}

\subsection{{Complexity of Nayak's}}    
\label{sec:complexity_of_nayak_scite_nandita2013}
{The optimization problem in Nayak's\cite{Nandita2013} is:
\begin{equation*}
    (\bD, \bS, \bW) = \arg\min_{\bD, \bS, \bW}\norm{\bY - \bD\bS}_F^2 + \lambda\norm{\bS}_1 + \norm{\bS - \bW\bY}_F^2.
\end{equation*}
$\bS$ is estimated via the gradient descent method that is an iterative method whose main computational task in each iteration is to calculate the gradient of $Q(\bS) = \norm{\bY - \bD\bS}_F^2 + \norm{\bS - \bW\bY}_F^2$ with respect to $\bS$. We have:
\begin{equation*}
  \label{eqn:gradnayak}
    \frac{\partial Q(\bS)}{\partial \bS} = 2\Big((\bD^{\top}\bD + \mathbf{I})\bS - (\bD^{\top} -\bW)\bY \Big).
\end{equation*}
where $\bD^{\top}\bD + \mathbf{I}$, and $(\bD^{\top} - \bW)\bY$ could be precomputed and at each step, only $(\bD^{\top}\bD + \mathbf{I})\bS$ need to be recalculated after $\bS$ is updated.
With $\bD \in \R^{d\times ck}, \bS \in \R^{ck \times cN}, \bY \in \R^{d \times cN}, \bW \in \R^{ck \times d}$, the complexity of the sparse coding step can be estimated as:
\begin{eqnarray}
\label{eqn:compnayak}
     T_{\text{Nayak's}} &\approx& (ck)d(ck) + 2(ck)d(cN) + 2q(ck)^2cN, \\
     &= &c^2kN(2d + 2qck) + c^2dk^2.
\end{eqnarray}
with $q$ being the average number of iterations needed for convergence.
Here we have ignored matrix subtractions, additions and scalar multiplications and focused on matrix multiplications only. We have also used the approximation that complexity of $\bA\bB$ is $2mnp$ where $\bA \in \R^{m\times n}, \bB \in \R^{n \times p}$. The first term in (\ref{eqn:compnayak}) is of $\bD^{\top}\bD + \mathbf{I}$ (note that this matrix is symmetric, then it needs only half of regular operations), the second term is of $(\bD^{\top} -\bW)\bY$ and the last one comes from $q$ times complexity of calculating $(\bD^{\top}\bD + \mathbf{I})\bS$.
}
\subsection{{Complexity of FDDL}} 
\label{sec:complexity_of_fddlcite_meng2011fddl}
{The sparse coding step in FDDL\cite{Meng2011FDDL} requires solving $c$ class-specific problems:
\begin{eqnarray*}
\label{eqn:findSFDDL}
    \bS_i = \arg\min_{\bS_i} \Big\{ \norm{\bY_i - \bD\bS_i}_F^2 + \norm{\bY_i - \bD_i\bS_i^i}_F^2 + \sum_{j = 1, j \neq i}^c\|\bD_j\bS_i^j\|_F^2 \\
    + \lambda_2\big\{\norm{\bS_i - \bM_i}_F^2 - \sum_{k=1}^c\|\bM_k - \bM\|_F^2 + \eta\|\bS_i\|_F^2 \big\}  + \lambda_1\norm{\bS_i}_1\Big\},
\end{eqnarray*}
with $\bD = [\bD_1, \dots, \bD_c], \bS_i^{\top} = [(\bS_i^1)^{\top}, \dots, (\bS_i^c)^{\top}]$, and $\bM_k = [\bm_k, \dots, \bm_k] \in \R^{ck \times N}, \bM = [\bm, \dots, \bm] \in \R^{ck \times N}$ where $\bm_k$ and $\bm$ are the mean vector of $\bS_i$ and $\bS = [\bS_1, \dots, \bS_c]$ respectively. The algorithm for solving this problem uses Iterative Projective Method\cite{rosasco2009iterative} whose complexity depends on computing gradient of six Frobineous-involved terms in the above optimization problem at each iteration.}
\\
\\ {For the first three terms, the gradient could be computed as:
  \begin{equation}
      \label{eqn:fddl_123}
      2(\bD^{\top}\bD)\bS_i - 2\bD^{\top}\bY_i + \bmt
      2(\bD_1^{\top}\bD_1)\bS_i^1 \\
      \vdots \\
      2(\bD_i^{\top}\bD_i)\bS_i^i - \bD_i^{\top}\bY_i \\
      \vdots \\
      2(\bD_c^{\top}\bD_c)\bS_i^c
       \emt,
  \end{equation}
  where $\bD^{\top}\bD, \and \bD^{\top}\bY_i$ could be precomputed with the total cost of $(ck)d(ck) + 2(ck)dN = cdk(2N + ck)$; $\bD_i^{\top}\bD_i, \and \bD_i^T\bY_i $ could be extracted from $\bD^{\top}\bD, \and \bD^{\top}\bY_i$ at no cost; at each iteration, cost of computing $(\bD^{\top}\bD)\bS_i$ is $2(ck)^2N$, each of $(\bD_j^{\top}\bD_j)\bS_i^j$ could be attained in the intermediate step of computing $(\bD^{\top}\bD)\bS_i$. Therefore, with $q$ iterations, the computational cost of (\ref{eqn:findSFDDL}) is:
  \begin{equation}
  \label{eqn:cost123}
  	cdk(2N + ck) + 2qc^2k^2N.       	
  \end{equation}}

{For the last three terms, we will prove that:
\begin{eqnarray}
\label{eqn:dev1}
	\frac{\partial}{\partial \bS_i}\|\bS_i - \bM_i\|_F^2 &=& 2(\bS_i - \bM_i),  \\
\label{eqn:dev2}
	\frac{\partial}{\partial \bS_i} \sum_{k=1}^c \|\bM_k - \bM\|_F^2 &=& 2(\bM_i - \bM), 	\\
\label{eqn:dev3}
	\frac{\partial}{\partial \bS_i} \eta \|\bS_i\|_F^2 &=& 2\eta \bS_i.
\end{eqnarray}
Indeed, let $\bE_{m,n}$ be a all-one matrix in $\R^{m\times n}$, one could easily verify that:
\begin{equation*}	
    \bM_k = \frac{1}{N} \bS_k \bE_{N,N}; \quad \bM = \frac{1}{cN}\bS \bE_{cN,N} = \frac{1}{cN}\sum_{i=1}^c \bS_i \bE_{N,N};
\end{equation*}
\begin{equation*}	
    \bE_{m,n}\bE_{n,p} = n\bE_{m,p}; \quad (\mathbf{I} - \frac{1}{N} \bE_{N,N}) (\mathbf{I} - \frac{1}{N} \bE_{N,N}) ^{\top} = (\mathbf{I} - \frac{1}{N} \bE_{N,N}).
\end{equation*}
Thus, (\ref{eqn:dev1}) can be obtained by:
\begin{multline*}
	\frac{\partial}{\partial \bS_i}\|\bS_i - \bM_i\|_F^2 = \frac{\partial}{\partial \bS_i}\|\bS_i - \frac{1}{N} \bS_i\bE_{N,N}\|_F^2\\
	= \frac{\partial}{\partial \bS_i} \|\bS_i(\mathbf{I} - \frac{1}{N} \bE_{N,N})\|_F^2  = 2\bS_i(\mathbf{I} - \frac{1}{N} \bE_{N,N}) (\mathbf{I} - \frac{1}{N} \bE_{N,N}) ^{\top} \\=  2\bS_i(\mathbf{I} - \frac{1}{N} \bE_{N,N}) = 2(\bS_i - \bM_i).
\end{multline*}}
{For (\ref{eqn:dev2}), with simple algebra, we can prove that:
\begin{equation*}
    \frac{\partial}{\partial \bS_i} \|\bM_i - \bM\|_F^2 = \frac{2(c-1)}{cN}(\bM_i - \bM)\bE_{N,N} = \frac{2(c-1)}{c}(\bM_i - \bM).
\end{equation*}
\begin{equation*}
     \frac{\partial}{\partial \bS_i} \|\bM_k - \bM\|_F^2 = \frac{2}{cN}(\bM - \bM_k)\bE_{N,N} = \frac{2}{c} (\bM - \bM_k), (k \neq i).
\end{equation*}}

\par {Compared to (\ref{eqn:fddl_123}), calculating (\ref{eqn:dev1}), (\ref{eqn:dev2}) and (\ref{eqn:dev3}) require much less computation. As a result, the total cost of solving $\bS_i$ approximately equals to (\ref{eqn:cost123}); and the total estimated cost of sparse coding step of FDDL is estimated as $c$ times cost of each class-specific problem and approximately equals to:
\begin{equation*}
T_{\text{FDDL}} \approx c^2dk(2N + ck) + 2qc^3k^2N = c^2kN(2d + 2qck) + c^3dk^2.
\end{equation*}
Final analyzed results of four different dictionary learning methods are reported in Table \ref{tab:complexity}.}
\bibliographystyle{IEEEtran}
\bibliography{TMI_Revised_arxiv}

\end{document}